\documentclass[sigconf]{acmart}

\usepackage{graphicx}
\usepackage[ruled, lined, linesnumbered, commentsnumbered, longend]{algorithm2e}
\usepackage{xcolor}
\usepackage{pifont}
\usepackage{makecell}
\usepackage{amsfonts}
\usepackage{booktabs}
\usepackage{multirow}
\usepackage{siunitx}
\usepackage{color, colortbl}
\definecolor{Gray}{gray}{0.9}

\usepackage{refcount}
\usepackage{amsfonts}
\usepackage{subfig}
\usepackage{caption}

\def\T{{\scriptscriptstyle\mathsf{T}}}
\DeclareMathOperator*{\argmin}{arg\,min}

\newcommand{\petgnnh}{HyperGene}

\theoremstyle{definition}
\newtheorem{definition}{Definition}

\newtheorem{pro-stat}{Problem Definition}

\newcommand{\hh}[1]{{\small\color{red}{\bf HH: #1}}}
\newcommand{\bx}[1]{{\small\color{blue}{\bf BX: #1}}}

\newcommand{\hide}[1]{}
\newcommand{\mkclean}{
	\renewcommand{\hh}[1]{}
	\renewcommand{\bx}[1]{}
}

\AtBeginDocument{%
  \providecommand\BibTeX{{%
    \normalfont B\kern-0.5em{\scshape i\kern-0.25em b}\kern-0.8em\TeX}}}

\renewcommand\footnotetextcopyrightpermission[1]{} 

\setcopyright{acmcopyright}
\copyrightyear{2021}
\acmYear{2021}

\begin{document}
\setlength{\abovedisplayskip}{1.4pt}
\setlength{\belowdisplayskip}{1.4pt}
\setlength{\abovedisplayshortskip}{1.4pt}
\setlength{\belowdisplayshortskip}{1.4pt}
\newcommand\mycommfont[1]{\small\ttfamily\textcolor{blue}{#1}}
\SetCommentSty{mycommfont}
\mkclean
%
\title{Hypergraph Pre-training with Graph Neural Networks}

%
\author{Boxin Du*, Changhe Yuan$\dagger$, Robert Barton$\dagger$, Tal Neiman$\dagger$, Hanghang Tong*} 
\affiliation{ 
  \institution{*Department of Computer Science, University of Illinois at Urbana-Champaign}
  \institution{$\dagger$Amazon, \{ychanghe, rab, talneim\}@amazon.com}
}
\email{*{boxindu2, lihuil2, htong}@illinois.edu }



\begin{abstract}

Despite the prevalence of hypergraphs in a variety of high-impact applications, there are relatively few works on hypergraph representation learning, most of which primarily focus on hyperlink prediction, and often restricted to the transductive learning setting. Among others, a major hurdle for effective hypergraph representation learning lies in the label scarcity of nodes and/or hyperedges. To address this issue, this paper presents an end-to-end, bi-level pre-training strategy with Graph Neural Networks for hypergraphs. The proposed framework named \petgnnh\ bears three distinctive advantages. First, it is capable of ingesting the labeling information when available, but more importantly, it is mainly designed in the self-supervised fashion which significantly broadens its applicability. Second, at the heart of the proposed \petgnnh\ are two carefully designed pretexts, one on the node level and the other on the hyperedge level, which enable us to encode both the local and the global context in a mutually complementary way. Third, the proposed framework can work in both transductive and inductive settings. When applying the two proposed pretexts in tandem, it can accelerate the adaptation of the knowledge from the pre-trained model to downstream applications in the transductive setting, thanks to the bi-level nature of the proposed method. The extensive experimental results demonstrate that: (1) \petgnnh\ achieves up to $5.69\%$ improvements in hyperedge classification, and (2) improves pre-training efficiency by up to $42.80\%$ on average. 

\end{abstract}

%
%


%

%

%
\maketitle
\section{Introduction}

Hypergraph, as a generalization of the regular graph data, is ubiquitous in various domains, and has drawn increasing attention recently \cite{vaida2019hypergraph, zhang2019hyper, zhang2018beyond}. Different from traditional regular \hh{let us call it regular graphs. plain graphs means graphs without attributes}graphs, which consist of nodes and edges to represent \textit{pairwise} relations between nodes, the hyperedges in the hypergraphs contain a collection of nodes, which represent \textit{high-order} relations. For example, in the clinical studies of the pharmacological mechanism \cite{sindrup1999efficacy, vaida2019hypergraph}, the effects of medical treatment is often the result of the combined interactions of a set of drugs. Here the combination of drugs for one disease could consist of one hyperedge. In the bioinformatics research, protein/multi-protein complexes, which consist of different collections of protein molecules, display different functions \cite{pinero2020disgenet}. Here a collection of protein molecules could be one hyperedge. In the social network domain, a group of users who participate in the same event could be an one hyperedge of that event \cite{yang2019revisiting, yadati2018link}. In the academic domain, authors of the same academic paper could be an hyperedge of the paper they jointly publish \cite{motl2015ctu}. 

\hide{
\begin{figure}[h]
    \centering
    \includegraphics[width=0.45\textwidth, height=0.28\textwidth]{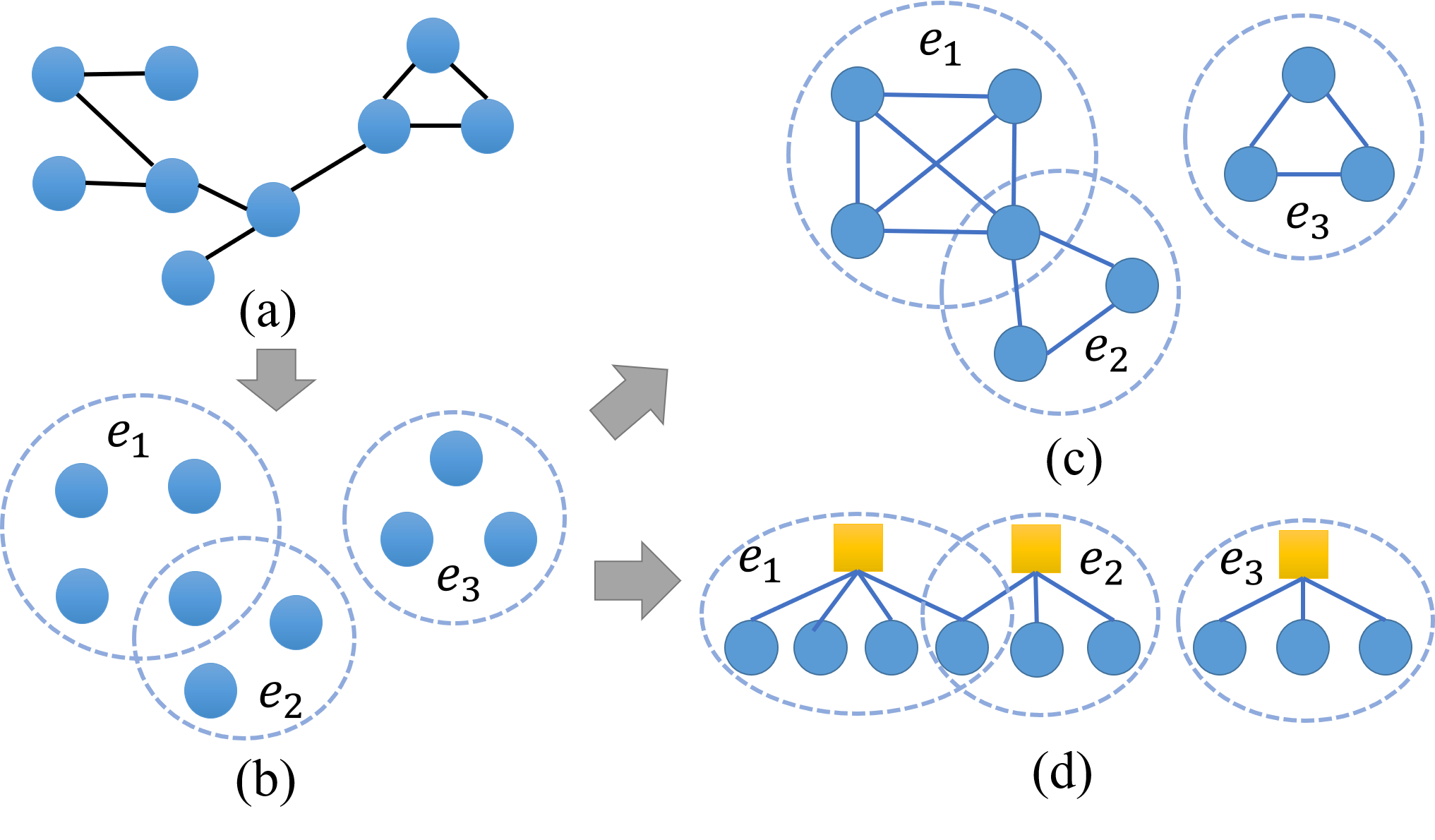}
    \caption{(a), (b): An example of plain graph and hypergraph; and (c), (d): two hyperedge expansion methods used in this paper. \bx{TODO: to be replaced by a real example of hypergraphs}}
    \label{fig:hypergraph}
\end{figure}
}

Representation learning on hypergraph offers a potentially promising way to streamline various hypergraph applications. However, the traditional representation learning methods on regular graphs are not directly applicable in capturing the {\em high-order} relations of the hypergraphs. To date, relatively few works on hypergraph representation learning exist, most of which primarily focus on hyperlink prediction (e.g. \cite{tu2017structural}, \cite{zhang2019hyper}, \cite{vaida2019hypergraph}, \cite{yadati2018link}). The major difficulties of representation learning on hypergraph are two-fold. First, the labels of diverse downstream tasks are usually very scarce, which makes it difficult for training the downstream neural models. Second, most of the recent hypergraph representation learning methods only work in the transductive learning setting (e.g. \cite{vaida2019hypergraph}, \cite{yadati2018link}, \cite{li2013link}, \cite{payne2019deep}, \cite{zhang2018beyond}, \cite{yadati2019hypergcn}). Specifically, these methods require all data to be seen during the training for feature generation or representation learning in the model, which renders the inability of these methods to handle unseen data. For example, in the hyperlink prediction problem, many methods require that all candidate hyperlinks to be seen during training for hyperlink prediction on unlabeled hyperlinks. 

\hide{
Second, many existing methods simplify the representation learning problem on hypergraph by transforming the original hypergraph into a plain graph of hypernodes or hyperedges, so that either traditional representation learning methods (e.g. unsupervised methods such as {\em DeepWalk}-based methods \cite{perozzi2014deepwalk} or supervised models such as GNN-based model \cite{hamilton2017inductive}) could be applied on the transformed graph, or structural features could be generated from it \cite{zhang2018beyond, vaida2019hypergraph, li2013link}. The problem of this method is that it fails to utilize both the node-level and hyperedge-level structural contexts of hypergraphs simultaneously. For example, conducting representation learning methods on the plain graph of nodes or its dual plain graph of hyperedges only captures local node/hyperedge structural characteristic. 
}

\hide{
Beside these two limitations, there commonly exists one blind-spot for the dataset. Although many current hypergraph datasets are constructed from plain graphs, the connections among nodes of the original plain graph should be strictly unavailable in the hypergraph scenario. For example, in Figure \ref{fig:hypergraph}, (a) represents the original plain graph, and (b) represents the hypergraph that is constructed by (a). For such constructed hypergraphs, the black links in (a) should never be known either explicitly (used for model) or implicitly (used for feature generation). 
}

In this paper, inspired by the recent advances of pre-training strategies developed in Natural Language Processing (NLP) \hh{spell it out the first time you use and then put 'NLP' in the parenthesis} community \cite{devlin2018bert} and Graph Neural Networks (GNN) \hh{spell it out}research (e.g. \cite{hu2019strategies}, \cite{jin2020self}, \cite{hu2020gpt}), we propose \petgnnh\ (i.e. a \underline{Hyper}graph pre-training framework with \underline{G}raph N\underline{e}ural \underline{Ne}tworks), and target on both inductive and transductive hyperedge classification. Compared with existing methods, our proposed method is characterized by the following three distinctive advantages \hh{the three advantages you put here do not align well with what we have in abstract. change it.}. First, the proposed pre-training framework is capable of leveraging labeled data (with supervised pre-training) as well as unlabeled data (with self-supervised pre-training), to learn transferrable knowledge for diverse downstream tasks without the help of extra domain-specific hypergraph datasets \cite{hu2019strategies, lu2021learning}. Second, our method explores bi-level (i.e. node-level and hyperedge-level) self-supervised pretext tasks, which aims at capturing the intrinsic {\em high-order} relationships of nodes and hyperedges respectively. Third, the proposed \petgnnh\ can work in both transductive and inductive settings. The pre-training strategy proposed for transductive setting is adaption-aware, in a sense that the pre-trained model could be more adaptive to the downstream tasks compared to traditional pre-training methods, and meanwhile more efficient.

\hide{
First, our pre-training strategy does not require extra domain-specific datasets, which often takes tremendous effort to establish for different domains of graphs \cite{hu2019strategies, lu2021learning}. Second, our method explores bi-level (i.e. node-level and hyperedge-level) self-supervised pretext tasks, for the purpose of capturing the intrinsic {\em high-order} relationships of nodes and hyperedges respectively. Third, our pre-training strategy is adaption-aware, in a sense that the pre-trained model could be more adaptive to the downstream tasks compared to traditional pre-training methods. Besides these distinctive advantages, our method is naturally applicable to inductive as well as transductive learning settings. The inductive learning setting does not require all data for heuristic feature or hidden representation generation in any training stage. With the pre-training framework of bi-level self-supervised pretext tasks, the pre-trained model could be generalized to unseen data. 
}

The main contributions of the paper are summarized as follows.

\begin{itemize}
    \item \textbf{Novel Framework}. We propose a bi-level pre-training framework for hypergraph representation learning named \petgnnh, equipped with two mutually complementary self-supervised pretext tasks. The proposed framework can be applied to both transductive and inductive settings.  For the tranductive setting, the proposed \petgnnh\ further embraces an adaptation-aware pre-training strategy to accelerate the knowledge transfer from the pre-trained model to the downstream applications.
    \item \textbf{Empirical Evaluations.} We perform extensive experiments to demonstrate the efficacy of \petgnnh. In particular, the proposed \petgnnh\ (1) outperforms {\em all} baselines across all datasets for inductive hyperedge classification, with an up to $5.69\%$ improvements over the best competitor\hh{over the best competitor?},  and (2) improves pre-training efficiency by up to $42.8\%$ on average. In addition, the proposed \petgnnh\ has been successfully applied to a real-world large online e-commerce application, namely inconsistent variation family (IVF) classification , outperforming the current baseline models therein.
\end{itemize}


The rest of the paper is organized as follows. Section \ref{problem-definition} formally describes the problem studied by the paper. Section \ref{method} presents the proposed model and Section \ref{experiments} shows the experiments on public datasets and the real-world case study is shown in Section \ref{case}. Related work is introduced in Section \ref{related-work}. The paper is finally concluded in Section \ref{conclusion}. 


\section{PROBLEM DEFINITION}\label{problem-definition}

The main notations used in this paper are summarized in Table \ref{tab:notations}. We first define the hypergraphs as follows. 

\begin{definition}{\textbf{Hypergraph:}}
\hh{(1) why N not n? (2) for the number of nodes in a given hyper edge, here you use k, but later you use n and what is worse, k is also used in eq4 (3) in table 1, you only have F, but here have Fn and Fe}A hypergraph is represented by $G = (\mathcal{V}, \mathcal{E}, \mathbf{F}^{(n)})$, in which $\mathcal{V} = \{ v_1, v_2, ..., v_n\}$ is the set of $n$ nodes and $\mathcal{E} = \{e_1, e_2, ..., e_m\}$ is a set of $m$ hyperedges. $e_i = \{v_j^{(i)}\}, 1 \leq j \leq n$ represents the $i$-th hyperedge in which the nodes $v_j^{(i)}\in \mathcal{V}$. We say that node $v_j$ is {\em inside} hyperedge $e_i$. $\mathbf{F}^{(n)}$ is the feature matrix\footnote{Optionally, there also might be feature matrix for hyperedges $\mathbf{F}^{(h)}$. } for nodes. 
\end{definition}
A hypergraph incidence matrix \cite{yadati2019hypergcn} $\mathbf{M}\in \mathbb{R}^{n\times m}$ is defined such that $\mathbf{M}(i, j)=1$ if node $i$ appears in hyperedge $j$ and otherwise $0$. From $\mathbf{M}$, we can build an adjacency matrix $\mathbf{A} = \mathbf{M}^{\T}\mathbf{M}$, in which $\mathbf{A}(i, j)$ indicates the number of nodes that appear in both hyperedge $i$ and hyperedge $j$. $\mathbf{A} = \mathbf{0}$ when hyperedges do not share nodes. 


\hh{let's bring preliminaries of gnn here: (1) you need gnn in def 2; and (2) it will make section 3 a bit shorter - right now, you have 7 subsections in section 3}

Before giving the definition of inductive hyperedge classification, we provide brief preliminaries on Graph Neural Networks.

\noindent \textbf{Preliminaries on Graph Neural Networks.} GNNs are powerful deep learning models on graphs. Representative models include Graph Convolutional Networks (GCN) \cite{kipf2016semi}, Graph Isomorphism Networks (GIN) \cite{xu2018powerful}, Graph Attention Networks (GAT) \cite{velivckovic2017graph}, etc. The intuition behind the GNN model is to learn the node representation by convolutionally aggregating both the node/edge features and the features of the node's local neighbors through neural networks. Message passing is often adopted as a convenient perspective to describe GNN models \cite{gilmer2017neural}. There are two steps in the message passing process, message passing and message updating. In message passing step, the node features are passed to its neighbors. In the message updating step, the received features are passed through an aggregation function (e.g., a neural network) for node representations. Typical message passing can be summarized as:

\begin{equation}\label{eq:passing}
    \mathbf{h}_{m_v^{t+1}} = P_t(\{\mathbf{h}_v^t, \mathbf{h}_w^t, \mathbf{e}_{vw}\}), \forall w \in \mathcal{N}(v)
\end{equation}

\begin{equation}
    \mathbf{h}_v^{t+1} = U_t(\mathbf{h}_v^t, \mathbf{h}_m{_v^{t+1}})
\end{equation}
$P_t$ and $U_t$ are the message passing function and node representation updating function of the $t-$th iteration respectively. $\mathbf{h}_v, \mathbf{h}_w$ are node representations (node $w$ is the neighbor of node $v$), which are initialized as node features. $\mathbf{e}_{vw}$ is the edge feature of edge between node $v, w$. Different GNN models differ in the functions $P_t()$ and $U_t()$. For example, GCN model \cite{kipf2016semi}, as one of the most representative GNN models, takes the summation of the neighboring nodes in the message passing step and attach a neural network module on the passed message and the node feature itself for feature aggregation. Recently there are also research on sampling strategies during message passing for removing redundant computations \cite{cong2020minimal}, so that the $w$ in Eq. \eqref{eq:passing} is does not always contain the entire neighborhood.

By using GNN as a neural function $f_{\Theta}(\cdot)$ for node representations, the inductive hyperedge classification is defined as follows. 

\begin{definition}{\textbf{Inductive Hyperedge classification:}}
Given a set of hyperedges $\mathcal{E} = \{e_1, e_2, ..., e_m\}$ which is not seen in the training stage, the goal of GNN model $f_\Theta(\cdot)$ is to learn embeddings for the downstream classifier\footnote{We will also use $g_{\Omega_i}$ to denote neural adjustment modules (e.g. using MLP) specified for pretext tasks with parameter $\Omega_i$ after GNN module.} $g_{\Omega}(\cdot)$ to classify them into $t$ categories. $g_{\Omega}(f_\Theta(e_i)) = \mathbf{p}_i, i \in \{1, 2, ...,  m\}$, and $\mathbf{p}_i$ is the prediction vector for $e_i$ with a non-zero entry indicating $e_i$'s category. 
\end{definition}
By adopting pre-training strategy, model $f_\Theta(\cdot)$ is first trained on pretext task(s). Note that there could be more than one pretext task. The fine-tuning module can be represented as $f_{\Theta:\Theta_0=\Theta'}(\cdot)$ given the pre-trained GNN module $f_{\Theta'}(\cdot)$. Generally speaking, pre-training $f_{\Theta'}(\cdot)$ could be either supervised if the labels for the pretext task are available, or unsupervised, such as the self-supervised methods. 

\hh{(1) F vs. Fn vs. Fe; (2) what is H? }
\vspace{-1.0\baselineskip}
{
	\begin{table}[H]
		\caption{Symbols and Definition\bx{TODO: check all notations}}
		\vspace{-1.0\baselineskip}
		\label{tab:notations}
		\centering
		\resizebox{0.5\textwidth}{!}{
			\begin{tabular}{|c|l|}\hline
				\textbf{Symbols} & \textbf{Definition} \\
				\hline
				$G=(\mathcal{V},\mathcal{E},\mathbf{F}^{(n)})$ & a hypergraph of node set $\mathcal{V}$, edge set $\mathcal{E}$, and feature $\mathbf{F}^{(n)}$\\
				$\mathbf{M}$ & the hypergraph incidence matrix \\
				$\mathbf{A}$ & the adjacency matrix of hyperedges inferred from $\mathbf{M}$\\
				$\Theta, \Omega_i$ & parameters of GNN module and adjustment modules\\
				$f_\Theta(\cdot)$ & GNN module with parameter $\Theta$\\
				$g_{\Omega_i}(\cdot)$ & neural adjustment module with parameter $\Omega_i$\\
				$f_{\Theta:\Theta_0=\Theta'}(\cdot)$ & a pre-trained GNN module with initialization $\Theta'$\\
				\hline
				$\mathbf{h}^{(e)}$ & the hidden representation of hyperedge $e$\\
				
				$\mathbf{h}^{(s)}$ & the hidden representation of seed node $s$\\
				$\mathbf{h}^{(c)}$ & the hidden representation of context $c$\\
				$\mathbf{h}^{(n)}$ & the hidden representation of negative example $n$\\
				\hline \end{tabular}
				}\end{table}
}

\section{Proposed Pre-training Framework}\label{method}

In this section we present the proposed \petgnnh\ framework. We first present the challenges and key ideas in Section \ref{sec:ideas}, and then the bi-level self-supervised pretext tasks in Section \ref{sec:node-level} and Section \ref{sec:he-level}. Section \ref{sec:ada} introduces an adaptation-aware pre-training strategy, followed by the overall \petgnnh\ framework architecture in Section \ref{sec:architecture} with analysis. \hh{we have many subsections in this section. have a small paragraph to talk about the organization of this section first}

\subsection{Challenges and Key Ideas}~\label{sec:ideas}

\begin{figure}[h]
    \centering
    \includegraphics[width=0.48\textwidth, height=0.18\textwidth]{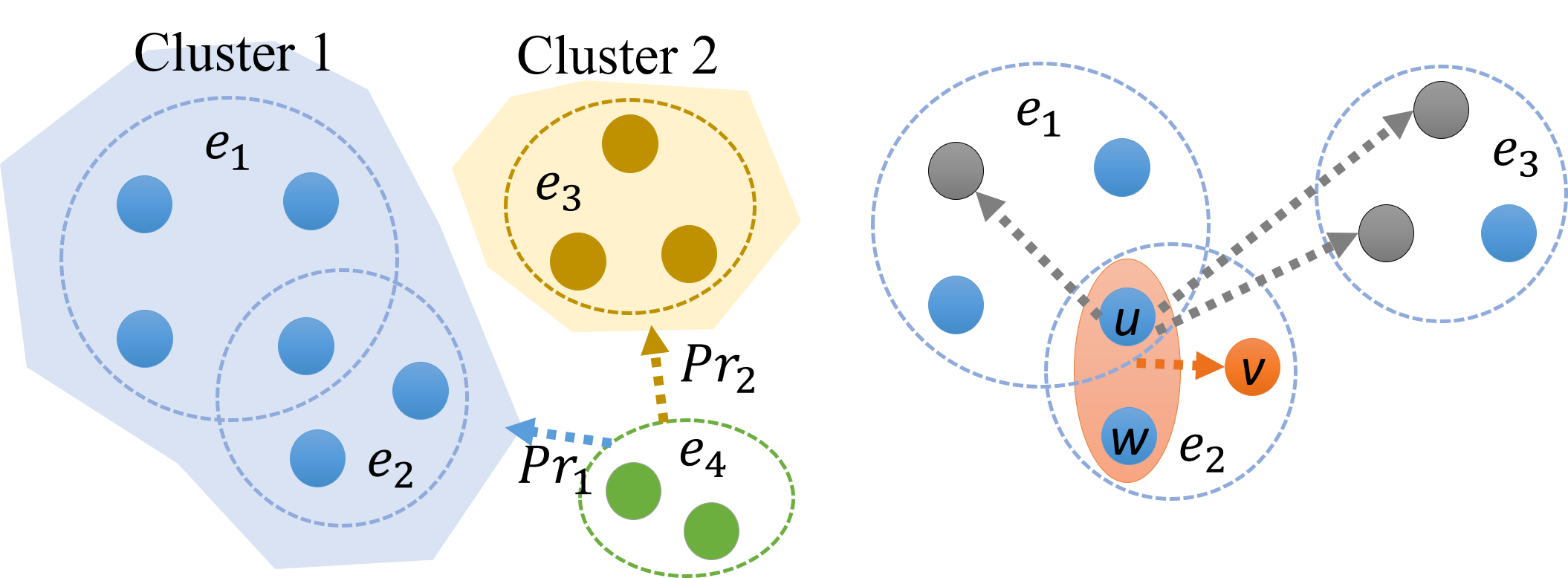}
    \vspace{-1.0\baselineskip}
    \caption{An illustrative example of the hyperedge-level pretext task (left), and the node-level pretext task (right). $Pr_1,Pr_2$ are probabilities for assigning $e_4$ to cluster 1 and cluster 2. The red area on the right subfigure shows the context of node $v$, and the gray nodes are the sampled negative examples of node $v$ and its context. Best viewed in color.}
    \label{fig:pretext}
\end{figure}

The first challenge for pre-training hypergraphs is how to design the self-supervised pretext tasks, since the {\em high-order} node relations of hypergraphs are significantly different from regular graphs structurally. Our idea is to incorporate both node-level and hyperedge-level pretext tasks, which aim at capturing local as well as global contextual pattern of hypergraphs. Locally, the node inside one specific hyperedge should be distinguished from nodes outside this hyperedge given the context of node. For one specific hyperedge and a given inside node, we define the context of the node as all other nodes inside the hyperedge (Figure \ref{fig:pretext}). Globally, the similarities between hyperedges ought to be preserved. However, calculating pairwise hyperedge similarities itself is challenging and costly (with at least $O(m^2)$ \hh{why?}time complexity for calculating every pair of hyperedges if the number of hyperedges is $m$). As an approximation for learning pair-wise hyperedge similarities, our idea is to first cluster the hyperedges, based on the features of nodes inside hyperedges or the hyperedge adjacency matrix when available, and then to preserve the membership characteristic of the hyperedge clusters. \hh{do we need this? can we remove it? again, let's focus on the main contributions of the paper. Note that the proposed self-supervised pretext tasks are performed on the same data as in downstream tasks, in order to learn the transferable general knowledge of the data, so that extra domain-specific data is not mandatory for pre-training. }

The second challenge is how to mitigate the divergence between the self-supervised pretext tasks and the downstream tasks. Even with two mutually complementary  pretext tasks, such divergence might still exist, in a sense that the well-trained pre-trained model might be overly fit on the pretext tasks, and could not optimally generalize on the downstream tasks. In the transductive setting, our idea is to design an adaption-aware pre-training strategy, which targets at learning a well-adaptive pre-trained model for downstream tasks. In this strategy, only one self-supervised pretext task (node-level) is fully trained on the  training data, and the other self-supervised pretext task (hyperedge-level) is applied on the unseen data for fast adaptation. 

\hide{
\noindent \textbf{B - Additional Issues.} \hh{let us move this part (the next two paragraphs) to appendix} Besides the above two major challenges, another problem is the absence of explicit, pairwise connections between nodes of hypergraphs, which makes GNN inapplicable. Our idea is to expand each hyperedge as a fully-connected graph (clique expansion). Intuitively, the nodes inside one hyperedge are closely related. Clique expansion allows the features to be passed to its close neighbors inside the same hyperedges during message passing step of GNN, which essentially filters feature noises and produces similar representations for contextually similar nodes \cite{nt2019revisiting}. 

In the case study, we demonstrate that the supervised pretext tasks could also be adopted for pre-training framework. Furthermore, the proposed self-supervised pretext task can be combined with supervised pretext tasks when pre-training the model. Note that for the supervised pretext tasks, it requires either domain knowledge or preliminary analysis to guarantee the positive correlation between the pretext tasks and the downstream tasks \cite{hu2019strategies}, otherwise the transfer from pretext tasks might impact the downstream tasks in a negative way. See details in Section \ref{case}. 
}

\vspace{-1.0\baselineskip}
\vspace{-1.0\baselineskip}
\begin{figure*}[t]
    \centering
    \includegraphics[width=0.97\textwidth, height=0.33\textwidth]{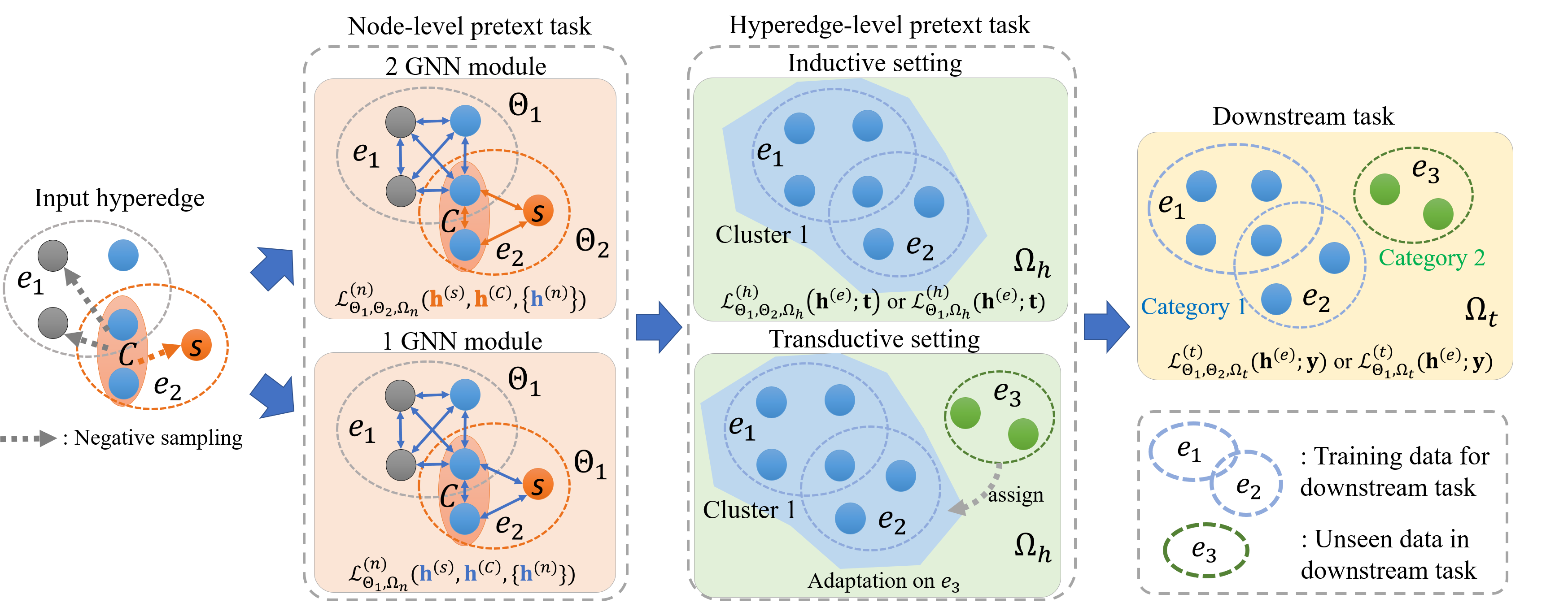}
    \vspace{-1.0\baselineskip}
    \caption{The pre-training and fine-tuning framework with node-level and hyperedge-level self-supervised pre-training/adaptation in \petgnnh\ for hyperedge classification. Best viewed in color.}
    \label{fig:model}
\end{figure*}

\subsection{Node-level Self-supervised Pretext Task}\label{sec:node-level}
\noindent \textbf{A - Task Description. }
In this pretext task, we aim at predicting the relationship between a given node and its hyperedge context (other nodes inside the hyperedge). Intuitively, we expect the node and the context share similar representation if they belong to the same hyperedge. 

To be specific, in the node-level self-supervised training, we first uniformly sample a seed node (e.g. $v^{(s)}_i$) inside each hyperedge, and obtain its corresponding context (e.g. $C_i$). The combined node-context pair ($v^{(s)}_i, C_i$) is a positive example. Next, for each pair ($v^{(s)}_i, C_i$), we adopt a negative sampling method for negative examples ($v^{(n)}_{ij} \sim Pr_{ij}$, the negative sampling probability) of the selected context. We utilize a GNN module inside the {\em clique-expansion} of hyperedges for the hidden representation of nodes. The representation of nodes corresponding to contexts are aggregated by a pooling layer for the context representations. The node-context relationship is learned via a binary cross-entropy objective\footnote{An objective variant is detailed in the appendix.}. For notation simplicity, let $\mathbf{h}$ be the representation learned by GNN module, and $\mathbf{\hat{h}} := g^{(n)}_{\Omega_n}(\mathbf{h})$ be the node representation after applying the neural adjustment function $g^{(n)}_{\Omega_n}(\cdot)$ for node-level pretext task \hh{this is odd, as we have not talked about the adaptation function yet}. 

\begin{equation}\label{eq:loss_node}
    \mathcal{L}^{(n)} = \sum_i\sum_j log[1 - \sigma((\mathbf{\hat{h}}_i^{(s)})^{\T}\mathbf{\hat{h}}_i^{(C)})] + log[\sigma((\mathbf{\hat{h}}_{ij}^{(n)})^{\T}\mathbf{\hat{h}}_i^{(C)})]
\end{equation}
where $\mathbf{\hat{h}}_i^{(s)}, \mathbf{\hat{h}}_i^{(C)}, \mathbf{\hat{h}}_{ij}^{(n)}$ are the hidden representation of seed node $i$, context of node $i$, and the negative sample $j$ of node $i$ after using the adjustment function, respectively. $\sigma(\cdot)$ is a sigmoid function. $(\mathbf{\hat{h}}_i^{(s)})^{\T}\mathbf{\hat{h}}_i^{(C)}$ is expected to be larger than $(\mathbf{\hat{h}}_{ij}^{(n)})^{\T}\mathbf{\hat{h}}_i^{(C)}$ since the positive seed node-context pair share similar feature distributions.

\noindent \textbf{B - Negative Sampling Method.} For a given pair of node and its corresponding context of one hyperedge, one naive negative sampling method is to uniformly sample nodes from outside the hyperedge. This is applicable even if the hyperedge adjacency matrix $\mathbf{A}$ is missing (zero), which is often the case in real-world applications (see Section \ref{case}). However, when such an adjacency matrix $\mathbf{A}$ can be obtained from the incidence matrix $\mathbf{M}$, this method does not distinguish between structurally close-by or distant hyperedges. Structurally close hyperedges tend to have very similar features or even same labels, and should be avoided as negative samples. We adopt the following negative sampling strategy such that the structurally close hyperedges would have exponentially lower probabilities to be sampled for negative nodes selection. 
For a given node $i$ the probability of sampling node $j\neq i$ is given as follows. \hh{k is double used in definition 1}
\begin{equation}\label{eq:prob}
    Pr_{ij} = \frac{exp(-\gamma \cdot\mathbf{\tilde{A}}_{ij}^k)}{\sum_{j}exp(-\gamma \cdot\mathbf{\tilde{A}}_{ij}^k)}
\end{equation}
in which $k, \gamma>0$ and $\gamma$ is a scaling scalar. $\mathbf{\tilde{A}}$ is the normalized adjacency matrix, and matrix $\mathbf{\tilde{A}}^k$ gives the number of $k-$length paths. We consider the nodes which are not reachable by $k-$length paths having high and equal probability of being sampled. 
This negative sampling method is referred to as the exponential sampling method (shorted as Exp.) in the rest of the paper. 

\subsection{Hyperedge-level Self-supervised Pretext Task}\label{sec:he-level}
In this pretext task, different from the local node-level pretext task, we strive to capture the more global patterns of the relationship of hypergraphs. As discussed in Section \ref{sec:ideas}, we aim at predicting hyperedges' cluster membership information. When the adjacency matrix of hyperedges $\mathbf{A}$ exists \hh{M for a hypergraph always exists, right?}, the clustering could be conducted on the regular graph of hyperedges. First the regular graph of hyperedges is constructed with adjacency matrix $\mathbf{A} = \mathbf{M}^{\T}\mathbf{M}$, such that $\mathbf{A}(i,j)$ is the number of nodes that exist in both hyperedge $i$ and $j$. Next, the METIS algorithm \cite{karypis1998fast} is applied to partition the graph of hyperedges into $q$ \hh{double used for context} clusters. $q$ is empirically selected, and is usually set larger or equal to the number of categories of hyperedges in the empirical experiments (see details in Section \ref{experiments}). Similar to the node-level self-supervised pretext task, the GNN module is applied inside the {\em clique expansion} of the hyperedges, and the representations for hyperedges are obtained by a graph pooling layer. Then by adopting the categorical cross-entropy loss, the hyperedge-level self-supervised task is written as: 

\hide{
 \bx{trying loss function that preserves the pair-wise distances or similarities: $\mathcal{L}(d(\mathbf{H}^{(\tau)},\mathbf{H}^{(\tau)}^{\T}, \mathbf{D}^{(\tau)}$) where $\tau$ is the index of batches}
\bx{The objective used is a regression objective:
\begin{equation}
    \mathcal{L}_h = ||f(\mathbf{H})^{\T}g^{(h)}_{\Omega_h}(\mathbf{H}) - \mathbf{A}^k||_F^2
\end{equation}
where $f, g$ are two adaption functions for the output embeddings of hyperedges. 
}
}

\begin{equation}\label{eq:loss_he}
    \mathcal{L}^{(h)} = - \sum_i [\log(\textrm{softmax}(g^{(h)}_{\Omega_h}(\mathbf{h}_i^{(e)}))]\circ \mathbf{y}^{(h)}_i]^{\T}\mathbf{1}
\end{equation}
where $\mathbf{h}_i^{(e)}$ is the hidden representation of the hyperedge $i$, $g^{(h)}_{\Omega_h}(\cdot)$ is a neural adjustment function to map the hyperedge representations to $q$-dimensional vector ($q$ is the number of clusters). $\mathbf{y}^{(h)}_i$ is the multi-class label vector of hyperedge $i$ indicating the cluster membership, and $\mathbf{1}$ is an all-one vector, for taking the summation of the $\log(\textrm{softmax}(\cdot))$ scores of the correct categories. 

\vspace{-1.5\baselineskip}
\subsection{Adaptation-aware Pre-training Strategy}\label{sec:ada} Traditional pre-training methods use a two-stage procedure, in which the first stage trains the pretext task until convergence with the self-labels, and the second stage trains the downstream task with the task labels. Specifically in our scenario with two self-supervised pretext tasks, the two-stage training can be realized in a serial procedure as follows.
\begin{subequations}
    \begin{equation}\label{eq:node-pre}
        \Theta'=\argmin_{\Theta}\mathcal{L}^{(n)}(g_{\Omega_n}(f_{\Theta}(\mathcal{E}_{train}, \mathbf{F}^{(n)}));\mathbf{y}^{(n)})
    \end{equation}
    \begin{equation}\label{eq:he-pre}
        \Theta^{(pre)} = \argmin_{\Theta}\mathcal{L}^{(h)}(g_{\Omega_h}(f_{\Theta:\Theta_0=\Theta'}(\mathcal{E}_{train}, \mathbf{F}^{(n)}));\mathbf{y}^{(h)})
    \end{equation}
    \begin{equation}
        \hat{\Theta} = \argmin_{\Theta}\mathcal{L}^{(t)}(g_{\Omega_t}(f_{\Theta:\Theta_0=\Theta^{(pre)}}(\mathcal{E}_{train}, \mathbf{F}^{(n)}));\mathbf{y}^{(t)})
    \end{equation}
\end{subequations}
where $\mathcal{L}^{(t)}$ and $\mathbf{y}_t$ are the loss function and labels for the downstream task respectively. The parameters of GNN module is first obtained by training the node-level pretext task and then by training the hyperedge-level pretext task. 

As discussed in Section \ref{sec:ideas}, the above strategy in the transductive setting might bring non-negligible divergence between pre-training and downstream task training, which would lead to a sub-optimal model. Moreover, since the hyperedge-level pretext task approximately preserves the pair-wise hyperedge distances, pre-training this task till convergence might result in a large divergence. To address this issue, we propose an adaption-aware pre-training strategy. The key idea is only performing node-level pre-training on the training set, and utilizing the hyperedge-level pretext task as adaption steps on the testing set (unseen when training the downstream task) in the transductive \hh{transductive?}setup. 

Specifically, we maintain Eq. \eqref{eq:node-pre}. But instead of training the hyperedge-level pretext task until convergence (Eq. \eqref{eq:he-pre}), we use it as an adaptation task, which can be represented as $s$ steps of gradient descent on testing hyperedges $\mathcal{E}_{test}$, so that Eq. \eqref{eq:he-pre} can be replaced as $s$ steps of: 
\begin{equation}\label{eq:ada_step}
    \Theta := \Theta - \epsilon \cdot \frac{\partial\mathcal{L}^{(h)}(g_{\Omega_h}(f_{\Theta:\Theta_0=\Theta'}(\mathcal{E}_{test}, \mathbf{F}^{(n)}));\mathbf{y}^{(h)})}{\partial \Theta}
\end{equation}
where $\epsilon$ is the learning rate. 
Compared with the traditional pre-training method, the advantages of the adaption-aware training strategy is two-fold. First, the pre-trained model for transferring general data knowledge is more adaptive to the downstream tasks. Second, it is a more efficient pre-training strategy, because only one pretext task needs to be fully trained, and the adaption (only a few steps) can be conducted whenever the testing data is available. 

\subsection{Proposed \petgnnh\ Framework }\label{sec:architecture}
The end-to-end model architecture is illustrated in Figure \ref{fig:model}, and the full algorithm is summarized in Algorithm \ref{alg:framework} (in Appendix). \hh{this part of writing is a bit confusing: 
The negative sampling strategy is first conducted for the input hyperedge set. Based on the actual set of hyperedges from which the negative sampling is conducted, two explorations in the experiments are detailed in the appendix. The node-level pretext task is adopted after negative sampling.}\bx{maybe change to in the footnote: 'In the Appendix, we elaborate two practical implementations of negative sampling, based on the actual set of hyperedges from which the negative sampling method is conducted.' or we can remove it}The negative sampling strategy is first conducted for the input hyperedge set, followed by the node-level pretext task\footnote{In the Appendix, we elaborate two practical implementations of negative sampling, based on the actual set of hyperedges from which the negative sampling method is conducted.}. Two variants of GNN modules are explored. First, the positive and negative hyperedges use different GNN modules for node/context representation (with parameters $\Theta_1$ and $\Theta_2$ in Fig. \ref{fig:model}). Second, contrarily \hh{what is this?} the positive and negative hyperedges share the parameters of the same GNN module. The proposed pre-training framework could support various types of GNN layers as the GNN module, such as GCN \cite{kipf2016semi}, GraphSAGE \cite{hamilton2017inductive}, GIN \cite{xu2018powerful}, etc. We use GIN \cite{xu2018powerful} module for clique expansion due to its superior empirical performance in our downstream task (see details in Section \ref{experiments}). After adopting the GNN module, a pooling layer is used for obtaining the context/hyperedge representation, such as mean pooling and set2set \cite{vinyals2015order}. In order to achieve permutation equivariance, set module could be adopted, such as Deep Sets \cite{zaheer2017deep}. Next, the hyperedge-level pretext task is adopted in two learning settings. In the inductive learning setting, the hyperedge-level pretext task is trained as an additional pre-training stage. If the node-level pretext task uses two GNN modules, the hyperedge representations are the concatenation of the outputs of the pooling layers in two GNN modules. Second, in the trasductive learning setting, in order to learn a pre-trained model which is adaptive to the downstream task, the hyperedge-level pretext task is used as adaptation stage as Eq.~\eqref{eq:ada_step}\hh{fill in}. The fine-tuning stage for the downstream task follows the pre-training, with the initialization of the pre-trained GNN module.

\noindent \textbf{A - Complexity Analysis.} For \petgnnh\ with uniform negative sampling method, the major computation of the model is applying GNN module on all hyperedges for training, which takes $O(d^2n'Lm \cdot iter)$ (e.g. for GCN module) where $d$ is the feature dimension, $n'$ is the number of nodes for each hyperedge, $L$ is the number of layers of GNN module, $m$ is the number of hyperedges, and $iter$ is the number of iterations. Here for notation simplicity, we assume all hyperedges share equal number of nodes $n'$, which is often much smaller than the number of hyperedges $m$ in a hypergraph. For \petgnnh\ with exponential negative sampling method, the major computation is calculating the $\mathbf{\tilde{A}}^k$ in Eq. \eqref{eq:prob}, which takes $O(kmn')$, where $m$ is the number of non-zero entries in $\mathbf{A}$. For METIS algorithm for hyperedge-level pretext task, its time complexity is $O(m+n'+q\cdot log(q))$, where $q$ is the number of clusters \cite{karypis1998fast}. 

\hh{move the remaining of this section to Appendix}

\noindent \textbf{B - Model Variants.}
Four model variants as well as additional experimental results are moved to the appendix for space. 

\hide{
\bx{Optional, or just move this paragraph to appendix}Here we briefly describe two other natural variants for this model. First, the representations of the sampled positive and negative seed nodes and context pair could be generated by different GNN module. Second, ranking loss functions (e.g. the cosine embedding loss) could be adopted instead of the binary classification loss, in order to learn the relative rankings for positive/negative node-context pairs. However, they both could not lead to significant performance improvement. See details in the appendix. 
}

\section{Experiments}\label{experiments}

In this section we present evaluation results of the effectiveness and efficiency of the proposed framework on public datasets. 

\subsection{Experimental Setup}
\noindent \textbf{A - Datasets.}
We use four public datasets to evaluate the proposed model. The statistics of the datasets are summarized in Table \ref{tab:datasets}.
\vspace{-0.5\baselineskip}
\begin{table}[ht]
    \centering
    \caption{The summary of datasets}
    \vspace{-1.0\baselineskip}
    \begin{tabular}{|c|p{13mm}|p{13mm}|p{14mm}|p{14mm}|}
        \hline
         Name & \# of nodes & \# of hyperedges & Min \# of nodes in hyperedge & Max \# of nodes in hyperedge \\ \hline
         \textit{Cora} & 2,708 & 2,427 & 2 & 169 \\ \hline
         \textit{Pubmed} & 19,717 & 3,887 & 3 & 20  \\ \hline
         \textit{Corum} & 6,132 & 4,736 & 2 & 131  \\ \hline
         \textit{Disgenet} & 8,352 & 8,386 & 3 & 487  \\ \hline
    \end{tabular}
    \label{tab:datasets}
\end{table}
\vspace{-1.0\baselineskip}
\begin{itemize}
\item \textit{Cora:} The Cora dataset consists of 2,708 scientific publications (nodes) classified into one of seven classes. The citation network consists of 5,429 edges between publications. The dataset contains text features for each publication, and it is described by a zero/one-valued word vector indicating the absence/presence of the corresponding word from the dictionary. The dictionary consists of 1,433 unique words \cite{motl2015ctu}. Note that the \textit{Cora} dataset is a regular graph dataset.

\item \textit{Pubmed:} The Pubmed Diabetes dataset consists of 19,717 scientific publications (nodes) from PubMed database pertaining to diabetes classified into one of three classes. The citation network consists of 44,338 links. This dataset contains text features for each publication which is described by a TF/IDF weighted word vector from a dictionary which consists of 500 unique words ~\cite{namata2012query}. The \textit{Pubmed} dataset is also a regular graph dataset. 

\item \textit{Corum \footnote{\url{https://mips.helmholtz-muenchen.de/corum/}}:} The Corum is the dataset of mammalian protein complexes. The dataset contains 6,132 types of proteins (nodes), and each protein complex consists of a collection of proteins. No direct connections between proteins in the complexes exist. The \textit{Corum} dataset is a hypergraph dataset. 

\item \textit{Disgenet \footnote{\url{https://www.disgenet.org/downloads}}:} The dataset contains 8,352 genes (nodes) and each disease (hyperedge) consist of a collection of genes. Each disease is classified into one of 23 MeSH codes. 21 of these codes are used as the hyperedge categories. Note that this dataset is highly unbalanced. 

\end{itemize}

\noindent \textbf{B - Experimental Setups.} We pre-process two versions of \textit{Cora} and \textit{Pubmed} datasets as \textit{Cora/Pubmed-clean/noisy}. The training details, hyper-parameter setting, dataset pre-processing and baseline adjustments are in Appendix for space. 

\subsection{Effectiveness Results on Public Datasets}



         
         
         

\begin{table*}
\caption{Accuracy of hyperedge classification (mean $\pm$ std in $\%$)}
\vspace{-1.0\baselineskip}
    \centering
    \begin{tabular}{l|p{18mm}|p{18mm}|p{19mm}|p{19mm}|p{18mm}|p{18mm}}
    \hline
         Models & Cora-noisy & Cora-clean & Pubmed-noisy & Pubmed-clean & Corum & Disgenet \\ \hline
         
         DHNE \cite{tu2017structural} & 69.85$\pm$1.01 & 72.48$\pm$0.52 & 78.65$\pm$1.59 & 83.23$\pm$0.74 & 53.11$\pm$1.39 &  30.35$\pm$0.83 \\ 
         
         Hyper-SAGNN \cite{zhang2019hyper} & 66.94$\pm$2.12 & 67.81$\pm$1.25 & 83.20$\pm$2.00 & 83.82$\pm$0.62 & 79.64$\pm$0.29 & 31.74$\pm$0.06  \\ 
         
         SAGE \cite{hamilton2017inductive} & 72.51$\pm$1.78 & 76.01$\pm$1.48 & 83.37$\pm$2.32 & 78.84$\pm$1.88 & 56.22$\pm$2.43 & 18.31$\pm$1.37  \\ 
         
         Joint Training & 70.84$\pm$0.67 & 81.65$\pm$1.16 & 85.39$\pm$1.48 & 86.45$\pm$0.98 & 76.85$\pm$0.77 & 34.02$\pm$1.28  \\ \hline
         
         DW \cite{perozzi2014deepwalk} & 73.39$\pm$1.64 & 81.66$\pm$1.41 & 82.73$\pm$1.70 & \underline{88.54$\pm$0.62} & 67.35$\pm$2.18 & 29.07$\pm$1.45 \\
         
         Deep-Hyperedge \cite{payne2019deep} & 74.35$\pm$0.64 & 72.66$\pm$0.93 & 64.33$\pm$0.71 & 81.58$\pm$1.01 & 82.74$\pm$1.64 & \textbf{35.46$\pm$1.08}  \\ \hline
         
         \rowcolor{Gray}
         \petgnnh\ (Uni., no Ada.) & \underline{77.98$\pm$1.81} & \textbf{86.41$\pm$1.62} & 85.18$\pm$0.68 & 88.21$\pm$1.51 & 84.07$\pm$0.70 & 33.23$\pm$1.45  \\
         
         \rowcolor{Gray}
         \petgnnh\ (Exp., no Ada.) & 77.68$\pm$2.67 & 81.74$\pm$1.14 & \underline{86.29$\pm$1.50} & 87.81$\pm$0.49 & \underline{84.49$\pm$0.47} & 34.10$\pm$0.83  \\
         
         \rowcolor{Gray}
         \petgnnh\ (Uni.) & 74.53$\pm$1.31 & \underline{83.86$\pm$3.55} & 85.52$\pm$0.95 & 87.23$\pm$0.96 & 84.31$\pm$0.99 & \underline{35.05$\pm$0.49}  \\ 
         
         \rowcolor{Gray}
         \petgnnh\ (Exp.) & \textbf{78.78$\pm$1.28} & 83.77$\pm$1.62 & \textbf{86.94$\pm$0.73} & \textbf{90.84$\pm$0.29} & \textbf{85.33$\pm$1.09} & 33.90$\pm$2.26  \\
         \hline
         
    \end{tabular}
    \label{tab:hyperedge}
\end{table*}
\begin{figure*}[h]
\hspace*{\fill}%
\begin{minipage}[h]{0.47\textwidth}
\centering
\vspace{0pt}
\includegraphics[width=1\textwidth, height=0.41\textwidth]{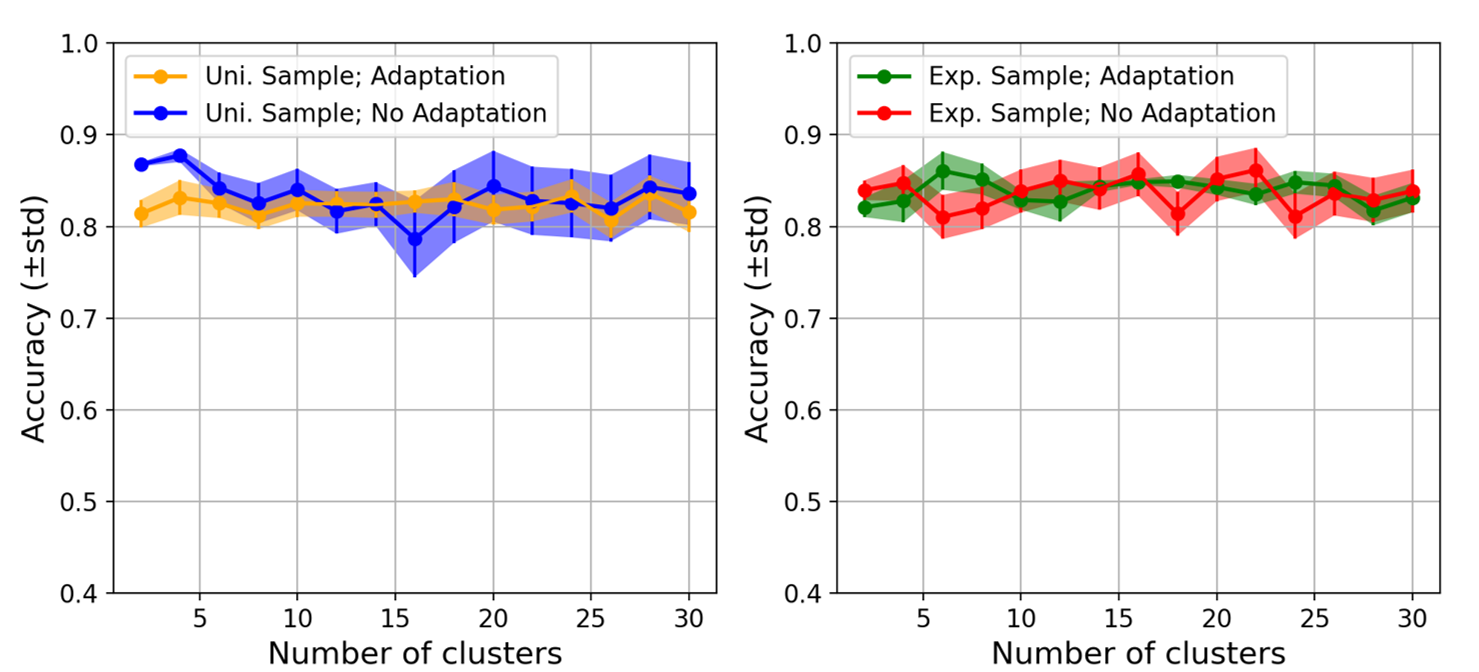}
\vspace{-2.0\baselineskip}
\caption{Accuracy ($\pm$std) vs. \# of clusters in Hyperedge-level pretext task on \textit{Cora-clean} dataset}
\label{fig:cora-sen}
\end{minipage}%
\hspace{8.00mm}
\begin{minipage}[h]{0.47\textwidth}
\centering
\vspace{0pt}
\includegraphics[width=1\textwidth, height=0.40\textwidth]{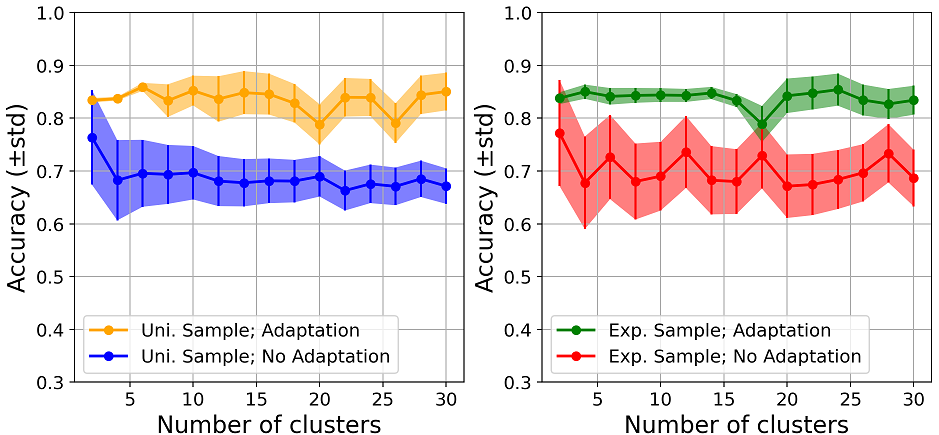}
\vspace{-2.0\baselineskip}
\caption{Accuracy ($\pm$std) vs. \# of clusters in Hyperedge-level pretext task on \textit{Corum} dataset}
\label{fig:protein-sen}
\end{minipage}%
\vspace{-1.0\baselineskip}
\end{figure*}
The results for hyperedge classification are presented in Table \ref{tab:hyperedge}. The metric is the multi-class classification accuracy. The results are mean and standard deviation values over ten runs. All supervised methods share the same training, validation, and testing ratio of 6:2:2. The best results are shown in bold fonts, and the second best results are shown with underlines. We adopt the two-GNN module for the inductive setting (no Ada. in Table \ref{tab:hyperedge}), and one-GNN module for the transductive setting (Ada. in Table \ref{tab:hyperedge}) because of overall better performance. Both exponential and uniform sampling are evaluated (shorted as Exp. and Uni. in Table \ref{tab:hyperedge}). Other results of model variants are detailed in the Appendix. For baselines, \textit{Hyper-SAGNN}, \textit{SAGE}, \textit{DHNE} and \textit{Joint Training} use inductive setting, and \textit{DW} and \textit{Deep-Hyperedge} use transductive setting. 

From the table, we can make the following observations. First, for inductive setting, \petgnnh\ outperforms all current baselines on all datasets (by up to $5.69\%$), and for transductive setting, \petgnnh\ outperforms all current baselines on five out of six datasets (by up to $4.75\%$). The framework with adaptation-aware pre-training outperforms the traditional pre-training method in five out of six datasets. The exponential sampling method is competitive with the uniform negative sampling method when no adaptation stage is used, but it achieves significant improvements when adaptation-aware pre-training is applied. The performance of the joint training model is competitive compared with the current baselines. In the current baselines, \textit{Deep Hyperedge}, \textit{Hyper-SAGNN} and \textit{DeepWalk} have relatively better performance over the rest of the baseline methods because \textit{DHNE} is originally for hyperlink prediction. Note that compared with \textit{Deep Hyperedge}, which requires hyperedge association, \petgnnh\ does not use such information in downstream tasks, but still outperforms \textit{Deep Hyperedge} in five datasets, and is also competitive on \textit{Disgenet}. For \textit{Disgenet} dataset, the relatively low accuracy is due to the highly imbalanced data with 21 hyperedge categories.

\subsection{Ablation Study}
In this section we conduct ablation study on the following setups. First, we compare the hyperedge classification performance by using different GNN modules in our framework. For all the different versions of GNN modules, we apply the adaptation-aware pre-training strategy, and the exact same hyper-parameters for the rest of the framework. The results are shown in Table \ref{tab:gnn_ablation}. The results are the mean and standard deviation values of ten runs. We can see that the GIN module overall shows the best performance over the rest of the GNN modules in our framework for the hyperedge classification task. Also, for the same GNN module, we can see that generally the exponential negative sampling method outperforms the uniform negative sampling method.

Second, we conduct the ablation study on the transductive hyperedge classification performance for bi-level self-supervised pretext tasks (Table \ref{tab:two_level_ablation}). We apply \petgnnh\ totally without pre-training stage (the first row), with only node-level self-supervised pretext task (the second row), and with only hyperedge-level self-supervised pretext task (the third row). We can observe that the \petgnnh\ framework with both two-level self-supervised pretext tasks shows the best performance, which demonstrates the effectiveness of both levels of self-supervised pretext tasks.

\subsection{Parameter Sensitivity Results}
Here we study the hyper-parameter sensitivity of our proposed framework. First, we show the results of the sensitivity of the number of clusters in the hyperedge-level pretext task in Fig. \ref{fig:cora-sen} and Fig. \ref{fig:protein-sen}. Note that we use the same model architecture with one-GNN module for all experiments in this subsection, whose setting is different from effectiveness experiments. The experiments are conducted in both inductive and transductive settings and with both uniform and exponential negative sampling methods. We observe that: (1) the framework shows relatively stable performance on a large range of the number of clusters; and (2) the framework that uses the exponential negative sampling method with adaptation-aware pre-training strategy overall shows stabler performance compared with the framework that uses uniform negative sampling, and does not use the adaptation-aware pre-training, which indicates better adaptation for the downstream task of the two-level adaption-aware pre-training framework. Second, the results of hyperedge classification accuracy vs. the number of adaptation steps are shown in Fig. \ref{fig:adastep-sen}. The results demonstrate slight performance drop but relatively stable over the tested adaptation steps in the range of $[1,20]$ for both exponential and uniform negative sampling method.
\begin{table*}
\caption{Accuracy of hyperedge classification for different GNN module (mean $\pm$ std in $\%$)}
\vspace{-1.0\baselineskip}
    \centering
    \begin{tabular}{l|p{20mm}|p{20mm}|p{20mm}|p{20mm}|p{20mm}|p{20mm}}
    \hline
         Models & Cora-noisy & Cora-clean & Pubmed-noisy & Pubmed-clean & Corum & Disgenet \\ \hline
         
         GCN (Uni.) & 75.58$\pm$0.98 & 82.09$\pm$1.26 & 83.59$\pm$0.63 & 85.96$\pm$2.27 & 65.23$\pm$0.48 & 32.36$\pm$0.44  \\
         GCN (Exp.) & \underline{76.88$\pm$2.08} & \textbf{84.65$\pm$3.32} & 84.14$\pm$1.62 & 86.06$\pm$1.94 & 83.34$\pm$0.25 & 32.84$\pm$0.18  \\ \hline
      
         SAGE (Uni.) & 75.28$\pm$0.69 & 80.68$\pm$0.94 & 83.89$\pm$1.27 & 86.15$\pm$2.71 & 64.98$\pm$0.45 & 33.33$\pm$0.79  \\ 
         
         SAGE (Exp.) & 75.52$\pm$1.01 & 81.04$\pm$2.09 & 85.18$\pm$0.86 & 86.94$\pm$0.32 & 66.03$\pm$0.28 & 31.69$\pm$0.21  \\ \hline
       
         GAT (Uni.) & 73.12$\pm$2.56 & 81.31$\pm$0.24 & 84.15$\pm$1.49 & 86.74$\pm$2.67 & 82.70$\pm$2.47 & 31.89$\pm$2.38  \\ 
         
         GAT (Exp.) & 71.21$\pm$0.72 & 80.15$\pm$1.68 & 84.79$\pm$0.72 & \underline{88.49$\pm$0.72} & \textbf{85.75$\pm$1.29} & 29.89$\pm$1.48  \\
         \hline
         
         \rowcolor{Gray}
         GIN (Uni.) & 74.53$\pm$1.81 & \underline{83.86$\pm$3.55} & \underline{85.52$\pm$0.95} & 87.23$\pm$0.96 & 84.31$\pm$0.99 & \textbf{35.05$\pm$0.49}  \\ 
         
         \rowcolor{Gray}
         GIN (Exp.) & \textbf{78.78$\pm$1.28} & 83.77$\pm$1.62 & \textbf{86.94$\pm$0.73} & \textbf{90.84$\pm$0.29} & \underline{85.33$\pm$1.09} & \underline{33.90$\pm$2.26}  \\
         \hline
         
    \end{tabular}
    \label{tab:gnn_ablation}
\end{table*}
\vspace{-1.0\baselineskip}
\begin{table*}
\caption{Ablation study on two-level self-supervised pretext tasks (mean $\pm$ std in $\%$)}
\vspace{-1.0\baselineskip}
    \centering
    \begin{tabular}{l|p{20mm}|p{20mm}|p{20mm}|p{20mm}|p{20mm}|p{20mm}}
    \hline
         Models & Cora-noisy & Cora-clean & Pubmed-noisy & Pubmed-clean & Corum & Disgenet \\ \hline
         
         No Pre-train & 69.12$\pm$0.48 & 81.48$\pm$1.24 & 84.40$\pm$1.53 & 84.60$\pm$2.49 & 78.25$\pm$0.13 & 32.76$\pm$1.54  \\
         
         Only Node & 74.23$\pm$0.84 & 82.27$\pm$1.25 & 85.04$\pm$0.37 & 86.84$\pm$1.94 & 82.52$\pm$2.14 & 33.75$\pm$1.70  \\ 
      
         Only Hyperedge & 73.24$\pm$1.53 & 82.78$\pm$1.30 & 86.29$\pm$1.36 & 85.67$\pm$0.38 & 84.17$\pm$1.56 & 33.40$\pm$0.65  \\ 
         
         \rowcolor{Gray}
         \petgnnh\ (Exp.) & \textbf{78.78$\pm$1.28} & \textbf{83.77$\pm$1.62} & \textbf{86.94$\pm$0.73} & \textbf{90.84$\pm$0.29} & \textbf{85.33$\pm$1.09} & \textbf{33.90$\pm$2.26}  \\ \hline
         
    \end{tabular}
    \label{tab:two_level_ablation}
\end{table*}
\hide{
\begin{figure}[h]
    \centering
    \includegraphics[width=0.45\textwidth, height=0.2\textwidth]{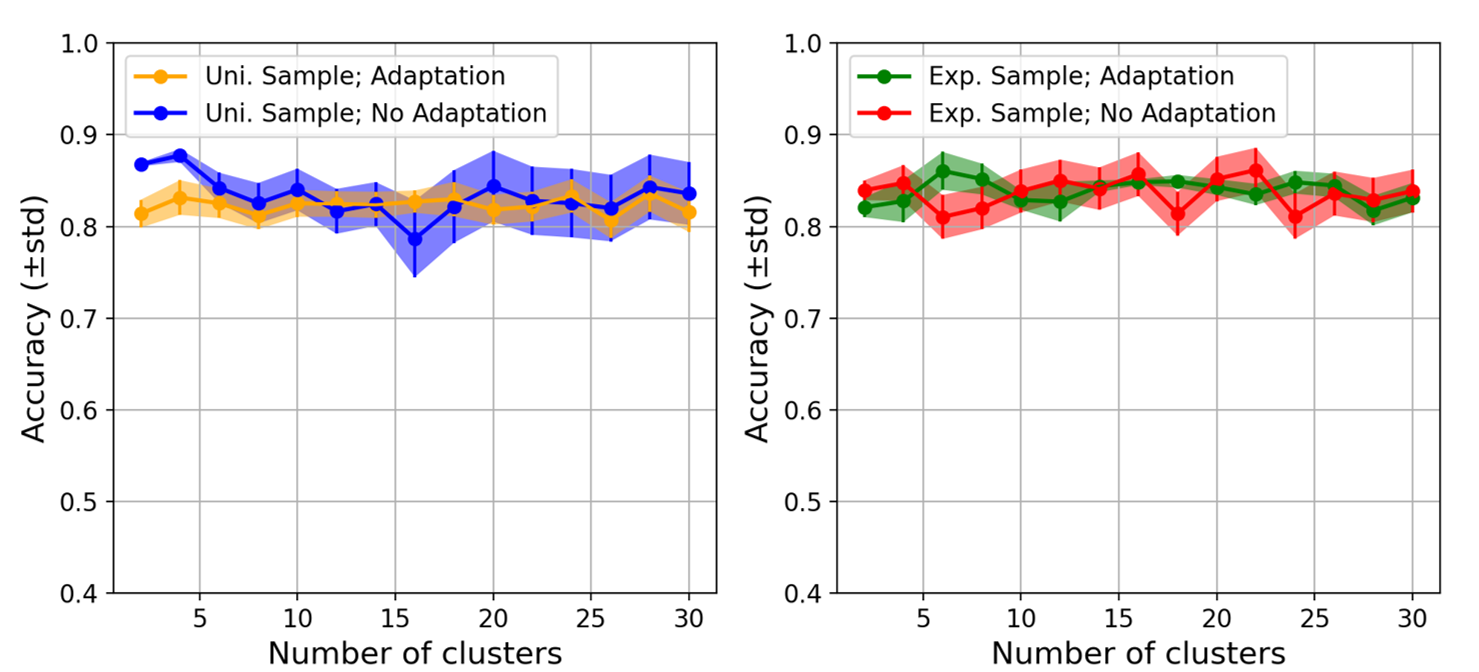}
    \vspace{-0.5\baselineskip}
    \caption{Accuracy ($\pm$std) vs. \# of clusters in Hyperedge-level pretext task on \textit{Cora-clean} dataset}
    \label{fig:cora-sen}
\end{figure}
\vspace{-1.5\baselineskip}
\begin{figure}[h]
    \centering
    \includegraphics[width=0.45\textwidth, height=0.2\textwidth]{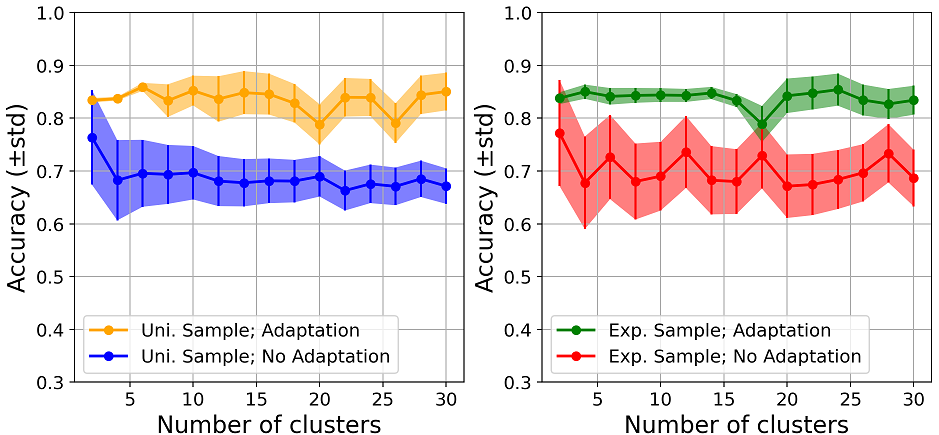}
    \vspace{-0.5\baselineskip}
    \caption{Accuracy ($\pm$std) vs. \# of clusters in Hyperedge-level pretext task on \textit{Corum} dataset}
    \label{fig:protein-sen}
\end{figure}
\vspace{-1.5\baselineskip}
}

\subsection{Efficiency Results}
Here we compare the efficiency of the adaptation-aware pre-training strategy with the traditional pre-training method as we describe in Subsection \ref{sec:ada}. The running time comparison of the pre-training stage is present in Fig. \ref{fig:efficiency}. For both methods, the number of node-level pre-training is set equal (till convergence). The number of adaptation steps for adaptation-aware pre-training method is set to 5 (equal to the setting in effectiveness evaluations). As we can see, adaptation-aware pre-training significantly reduces the pre-training time (by $38.2\%$ with Exp. sampling and $42.8\%$ with Uni. sampling). Besides, as we adopt local sampling in the uniform negative sampling method, the running time for uniform negative sampling achieves $\sim$7 times reduction compared with exponential negative sampling (see Appendix-A for more details). 
\begin{figure*}[h]
\hspace*{\fill}%
\begin{minipage}[h]{0.47\textwidth}
\centering
\vspace{0pt}
\includegraphics[width=1\textwidth, height=0.40\textwidth]{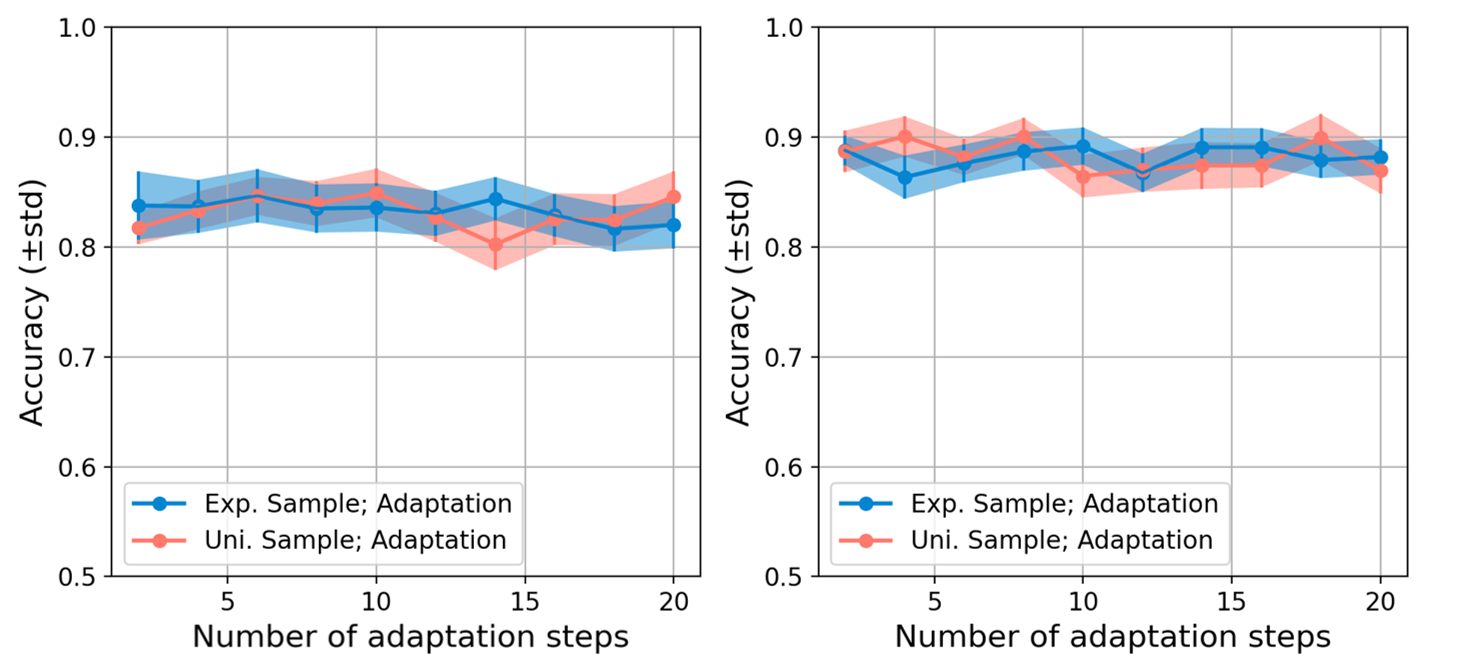}
\vspace{-2.0\baselineskip}
\caption{Accuracy ($\pm$std) vs. \# of adaptation steps on \textit{Cora-clean} (left) and \textit{Pubmed-clean} (right) dataset}
\label{fig:adastep-sen}
\end{minipage}%
\hspace{8.00mm}
\begin{minipage}[h]{0.47\textwidth}
\centering
\vspace{0pt}
\includegraphics[width=1\textwidth, height=0.40\textwidth]{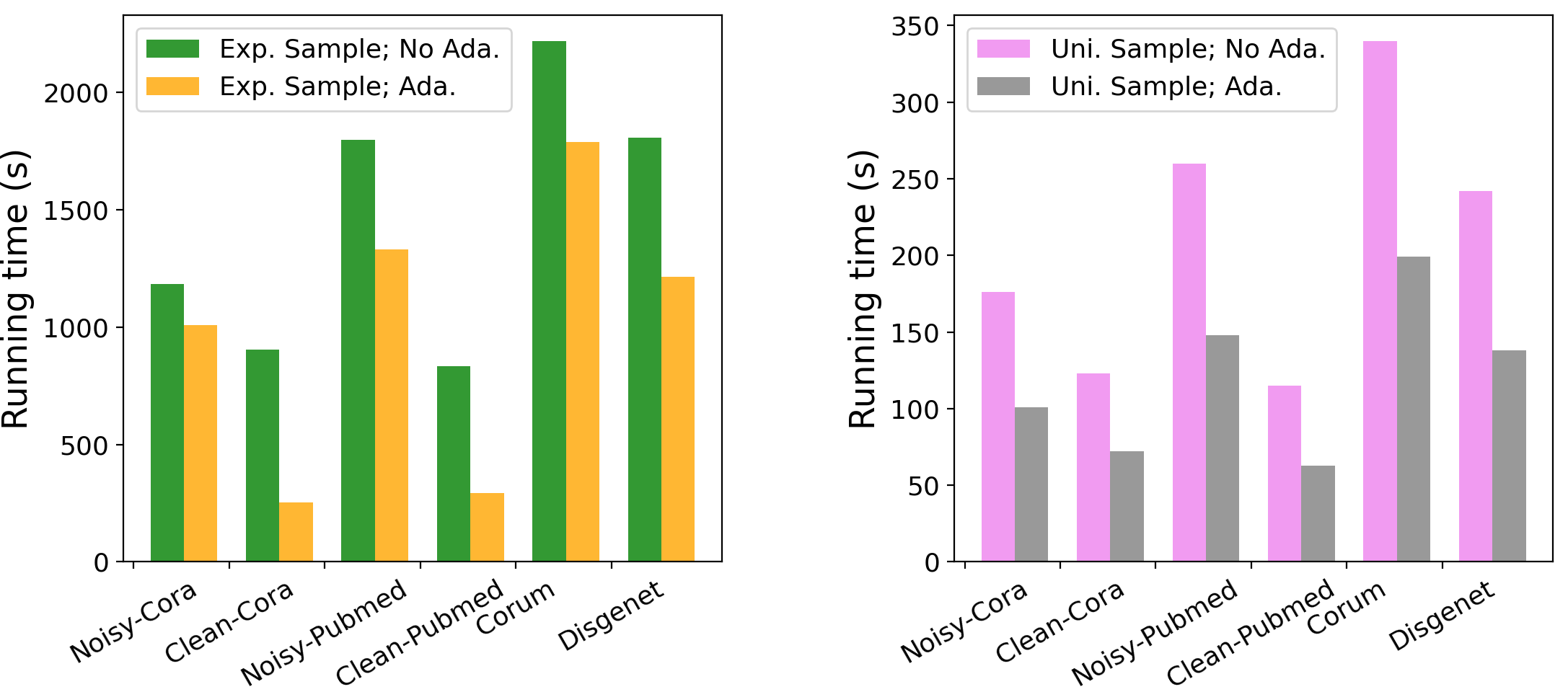}
\vspace{-2.0\baselineskip}
\caption{Running time comparison of the pre-training stage for traditional and adaptation-aware pre-training stragety.}
\label{fig:efficiency}
\end{minipage}%
\vspace{-1.0\baselineskip}
\end{figure*}

\hide{
\begin{figure}[h]
    \centering
    \includegraphics[width=0.48\textwidth, height=0.17\textwidth]{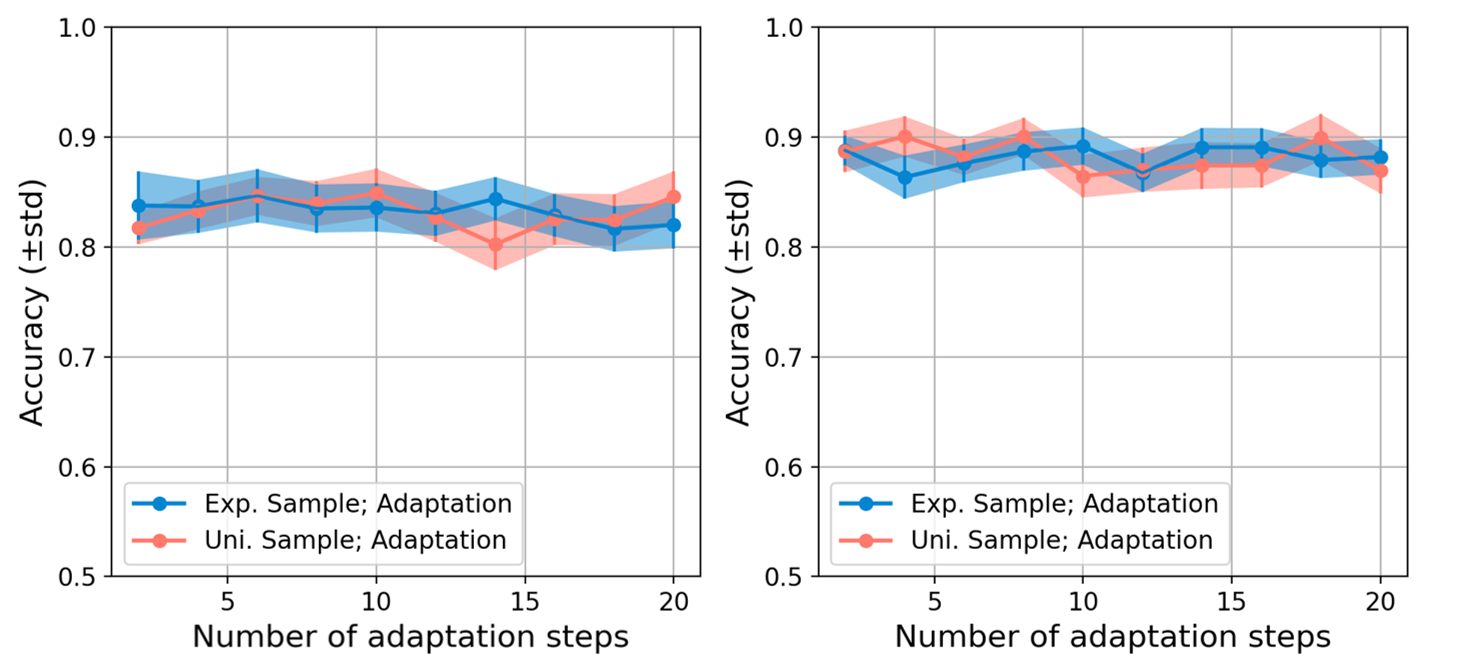}
    \vspace{-1.0\baselineskip}
    \caption{Accuracy ($\pm$std) vs. \# of adaptation steps on \textit{Cora-clean} (left) and \textit{Pubmed-clean} (right) dataset}
    \label{fig:adastep-sen}
\end{figure}
\vspace{-1.5\baselineskip}
\begin{figure}[h]
    \centering
    \includegraphics[width=0.48\textwidth, height=0.19\textwidth]{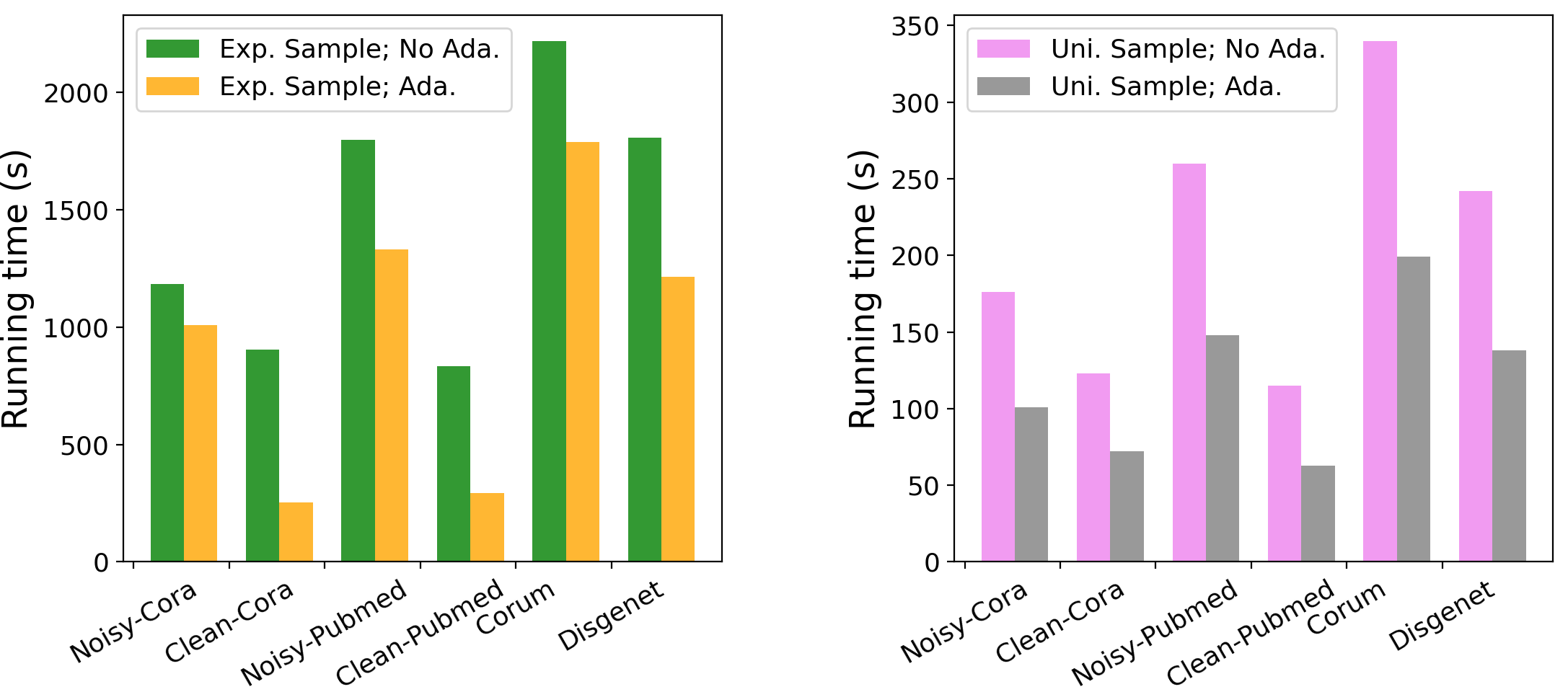}
    \vspace{-1.0\baselineskip}
    \caption{Running time comparison of the pre-training stage for traditional and adaptation-aware pre-training stragety.}
    \label{fig:efficiency}
\end{figure}
\vspace{-1.0\baselineskip}
}

\vspace{-1.0\baselineskip}
\section{A Case Study} \label{case}
In this section, we introduce a real-world application of the proposed model on the inconsistent variation family detection problem in a large online store, and show that how the proposed pre-training strategy can help to improve base model (without pre-training) as well as baselines, even when direct hyperedge associations are not available (the adjacency matrix $\mathbf{A}$ of hyperedges is a zero matrix).

\subsection{Preliminaries and Experimental Setup}
\noindent \textbf{A - Preliminaries} Large E-commerce services such as Amazon, Etsy, and eBay often utilize merchandise relationships such as variations and substitutes to improve their catalog quality. These relationships between merchandise items are the cornerstone of research in the field of relationship science. 

\begin{figure}[h]
    \centering
    \includegraphics[width=0.40\textwidth, height=0.23\textwidth]{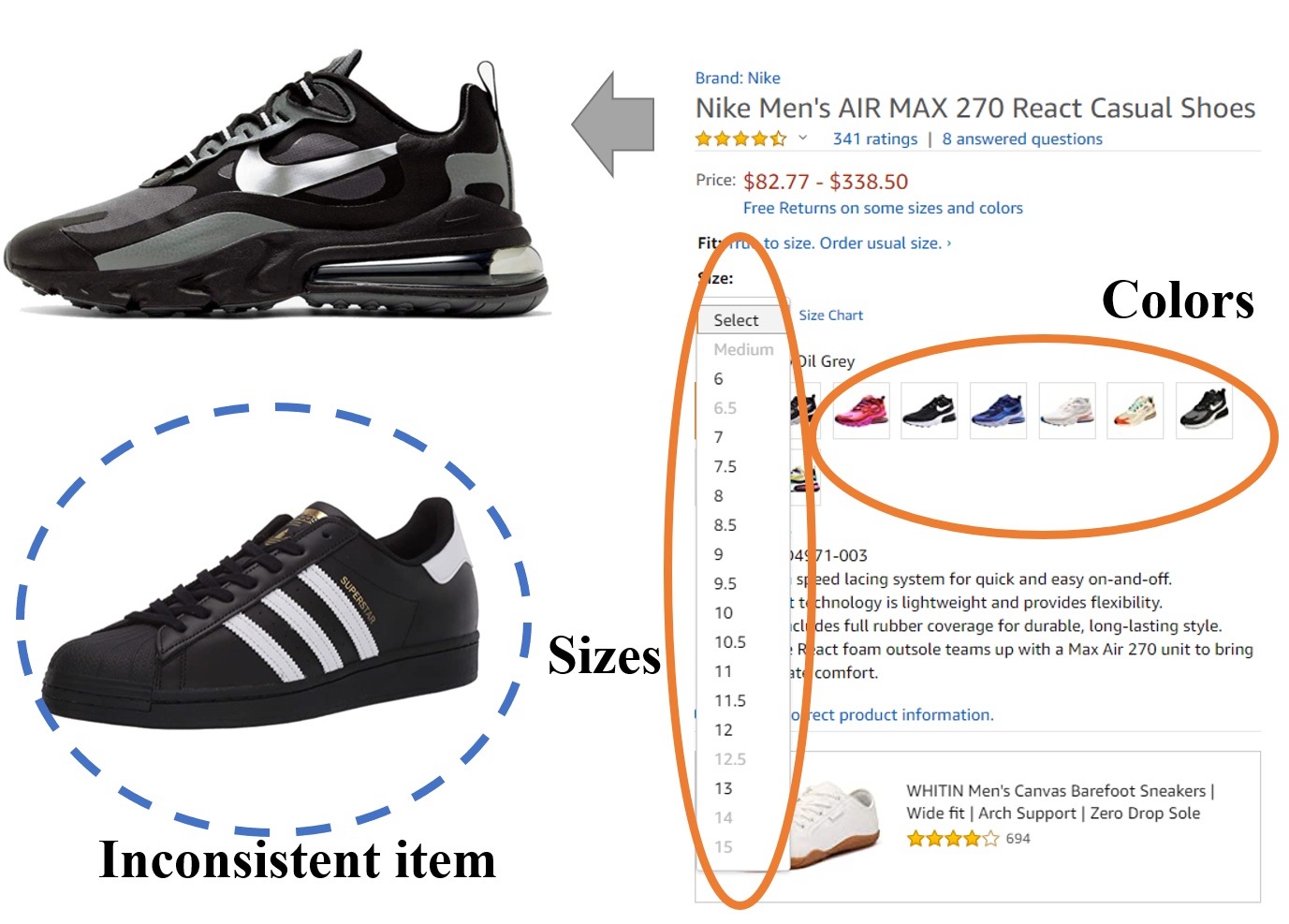}
    \vspace{-1.0\baselineskip}
    \caption{An example of correctly grouped variation family with variation theme attributes, and an inconsistent item.}
    \label{fig:family_example}
\end{figure}

Variation family refers to a family of product items which are functionally the same but differ in specific attributes. Such variation families are present in the same detail page. For example, in Fig. \ref{fig:family_example} the correctly grouped 'Nike Air Max 270 React' shoe family is shown in one detail page. All items inside this family are this particular model but have different sizes and colors. The grouping attributes of a variation family are shared by all items inside the variation family (e.g., brand 'Nike'), and the variation theme attributes are attributes which differ from item to item inside the variation family (e.g., size and color). When variation families contain inconsistent items (e.g., an Adidas shoe is grouped into the variation family in Fig. \ref{fig:family_example}), they are called inconsistent variation family (IVF). One of the most important task in the catalog system is the IVF detection. 

In order to solve this problem, each variation family can be seen as a hyperedge, in which every item should belong to the same merchandise if it is a consistent family. Then this specific problem is transformed into hyperedge classification problem. As for the pretext tasks, we adopt a task called variation theme learning, which is defined as:
\begin{definition}{\textbf{Variation Theme Learning:}}
Given a set of variation families (hypergraph $G=(\mathcal{V},\mathcal{E}, \mathbf{F}^{(n)})$ with $\mathcal{E} = \{e_1, ..., e_m\}$), the Variation Theme Learning aims at learning the variation theme attributes and grouping attributes of the families (i.e. which rows of $\mathbf{F}^{(n)}$ are grouping/variation theme attributes).
\end{definition}
\begin{table*}
\caption{Results of IVF classification for case study (mean $\pm$ std in $\%$)}
\vspace{-1.0\baselineskip}
    \centering
    \begin{tabular}{l|p{14mm}p{14mm}|p{14mm}p{14mm}|p{14mm}p{14mm}}
    \hline
         Dataset & \multicolumn{2}{c}{Category 1} & \multicolumn{2}{c}{Category 2} & \multicolumn{2}{c}{Category 3}\\ \hline
         Metric & AUC-I & AUC-C & AUC-I & AUC-C & AUC-I & AUC-C \\ \hline
         
         Hyper-SAGNN & 56.12$\pm$1.19 & 88.87$\pm$0.93 & 58.72$\pm$1.94 & 89.81$\pm$1.64 & 70.26$\pm$1.12 & 88.47$\pm$0.78 \\ 
         
         Hyper-SAGNN, pre-train & 60.26$\pm$1.42 & 90.12$\pm$0.38 & \underline{59.82$\pm$0.55} & 89.57$\pm$0.21 & 72.17$\pm$0.34 & 90.17$\pm$1.42 \\
         
         \petgnnh, no pre-train & 60.51$\pm$0.89 & 90.32$\pm$0.88 & 58.09$\pm$0.57 & 89.64$\pm$0.79 & 71.50$\pm$1.49 & 89.80$\pm$1.72 \\
         
         Joint Training & \underline{62.80 $\pm$1.33} & \textbf{90.99$\pm$0.78} & 56.73$\pm$1.41 & 89.55$\pm$0.82 & 71.50$\pm$1.52 & 89.80$\pm$1.39 \\ \hline
         
         \rowcolor{Gray}
         \petgnnh, Sup. pre-train & \textbf{64.06$\pm$1.27} & \underline{90.74$\pm$1.40} & 59.42$\pm$1.78 & \underline{89.84$\pm$0.70} & \underline{74.32$\pm$0.81} & \textbf{90.62$\pm$1.53}  \\ 
         
         \rowcolor{Gray}
         \petgnnh, Sup. pre-train $+$ Ada. & 61.86$\pm$1.32 & 90.31$\pm$1.80 & \textbf{61.73$\pm$1.24} & \textbf{90.42$\pm$1.74} & \textbf{75.42$\pm$2.01} & \underline{90.32$\pm$1.85} \\ \hline
    \end{tabular}
    \label{tab:multi-task}
\end{table*}

Note that variation families are independent, which makes the direct hyperedge associations unavailable. Most baselines from Section \ref{experiments} become inapplicable, such as \textit{SAGE}, \textit{DW}, \textit{Deep-Hyperedge}, and \textit{DHNE}. Our idea is to transform this problem as a hyperedge multi-label classification problem since variation families (hyperedges) might have multiple grouping/variation theme attributes. Due to the fact that the labels of grouping/variation theme attributes are available in our catalog system for variation family datasets, this pretext task in the pre-training stage could be supervised. During training, the hyperedge representation produced by \petgnnh\ is first used for the multi-label supervised pretext task (variation theme learning), and then the model is fine-tuned by the IVF detection task\footnote{As suggested by \cite{hu2019strategies}, through preliminary studies, these two tasks are highly likely to be positively correlated. Briefly, inconsistent variation families contain inconsistent items with different distributions of grouping attributes compared with other items, and vice versa. }. IVF detection task has binary labels indicating consistent and inconsistent families (binary classification).  


\noindent \textbf{B - Experimental Setup} We sample subsets of three product lines from the catalog system, named Category 1, Category 2, and Category 3. Some details of these datasets are show in Table \ref{tab:case_data}. 
\begin{table}[h]
    \centering
    \caption{Statistics of Datasets for Case Study}
    \vspace{-1.0\baselineskip}
    \begin{tabular}{|c|c|c|c|}
    \hline
         Name & \# of families & \# of items & \# of unique items \\ \hline
         Category 1 & 1,186 & 64,752 & 9,540 \\ \hline
         Category 2 & 2,169 & 48,860 & 8,474 \\ \hline
         Category 3 & 4,247 & 138,662 & 21,131\\ \hline
    \end{tabular}
    \label{tab:case_data}
\end{table}
The proposed method we select for this experiments is \petgnnh\ with supervised pre-training task (variation theme learning), and \petgnnh\ with supervised pre-training task plus node-level adaptation. For comparison, we use three strong baselines, including the joint training method of the pretext task (variation theme learning) and the targeted IVF classification using \petgnnh\ (denoted as 'Joint training' in Table \ref{tab:multi-task}), \petgnnh\ model without pre-training stage, and the \textit{Hyper-SAGNN} model with and without pre-training strategy. Note that the original \textit{Hyper-SAGNN} is not a pre-training model. We modify the model to adapt it in our pre-training framework as a strong baseline. Besides the above adaptation, since there are text features for all items in the variation family, we apply the pre-trained \textit{Glove} embeddings \cite{pennington2014glove} as features for items. 

The metric for this experiment is the PR-AUC for predicting inconsistent variation family (IVF) and consistent variation family (CVF), denoted as AUC-I and AUC-C respectively in Table \ref{tab:multi-task}. The results are the average of five runs. 


\subsection{Results on Datasets of Product Lines}

As shown in Table \ref{tab:multi-task}, first we can see that by using the supervised pretext task, the model performances are consistently better than those without pre-trained pretext task in all product lines. Second, for both methods using pretext task, the pre-training strategy outperforms the joint training strategy by AUC-I metric and they are very close by AUC-C metric. The \textit{Hyper-SAGNN} with the adapted pre-training stage also has relatively competitive performance compared with the proposed model. This indicates that in this real-world application of hyperedge classification, pre-training can improve the base model as well as baseline method, and pre-training is generally a better training strategy than joint training. Third, by utilizing the adaptation stage with the node-level self-supervised pretext task (the last row), the model is able to further improve the AUC-I performance on Categoty-2 and Category-3 product lines, which demonstrates the effectiveness of the adaptation-aware pre-training of \petgnnh\ in this application scenario. 
\vspace{-1.0\baselineskip}
\section{Related work}\label{related-work}
The related work of the proposed model can be roughly divided into three categories, the GNN-related research, the recent pre-training methods, and the hypergraph mining. 

\noindent \textbf{A - GNN-related Research.} GNN-related works are partially covered in Section \ref{problem-definition}. It becomes popular since Bronstein et al. introduce geometric deep learning \cite{bronstein2017geometric} to bring the convolutional network to the graph mining field, and Kipf et al. significantly simplify the graph convolutional process and develop GCN \cite{kipf2016semi}. To name a few representative works, Velivckovic et al. \cite{velivckovic2017graph} improve GCN by introducing attentions of neighborhoods, which is calculated dynamically at message passing step. Xu et al. study the expressiveness of GNN model compared with the closely related Weisfeiler-Lehman graph isomorphism testing algorithm, and develop Graph Isomorphism Networks \cite{xu2018powerful}. This direction still has numerous challenges.

\noindent \textbf{B - Pre-training Strategy.} After the pre-training strategy is first used in NLP communities \cite{devlin2018bert}, researches also begin to appear in the graph mining field. Hu et al. develop a new strategy and self-supervised methods for pretraining Graph Neural Networks \cite{hu2019strategies}. The key to the success of this method is to pretrain an expressive GNN at the level of individual nodes as well as entire graphs so that the GNN can learn useful local and global representations simultaneously. Most recently, Hu et al. \cite{hu2020gpt} propose a generative pre-training strategy for GNN models in which a self-supervised attributed graph generation task is introduced. Qiu et al. \cite{qiu2020gcc} propose a graph contrastive coding method for the pre-training stage in the GNN model. Jin et al. \cite{jin2020self} summarize and compare commonly used self-supervised training strategies on graphs and point out future directions. However, the pre-training strategy on graph/hypergraph data is still an open question with lots of potential for breakthrough in the immediate future. 

\noindent \textbf{C - Hypergraph Mining.} The hypergraph mining studies the combined relation pattern of inner-related items. Among some earliest work, Li et al. \cite{li2013link} transform the hyperlink prediction method \cite{jeh2002simrank} for ranking the hyperedges. Zhang et al. \cite{zhang2018beyond} use a coordinated matrix minimization method for nonnegative matrix factorization in the adjacency space of the hypergraphs for hyperedge prediction. More recently, representation learning and GNN based methods are developed for this direction. Tu et al. propose DHNE \cite{tu2017structural} which applies auto-encoder and deep neural networks for the representation learning of hyperedges. Hyper-SAGNN by Zhang et al. \cite{zhang2019hyper} combines the attention-based dynamic representations and the MLP-based static representations for hyperedge prediction. HGNN by Feng et al. \cite{feng2019hypergraph} generalizes the convolution operation to the hypergraph by hypergraph Laplacian. Deep Hyperedges \cite{payne2019deep} by Payne jointly uses contextual and permutation-invariant node memberships of hyperedges for hyperedge classification, which is relevant to the focus of this paper. DHGNN by Jiang et al. \cite{jiang2019dynamic} adopts vertex convolution and hyperedge convolution in the dynamic hypergraph representation learning. Yu et al. \cite{yu2021self} propose a multi-channel hypergraph convolutional network for social recommendation. Other recent neural hyperlink prediction methods include \cite{yang2019revisiting, yoon2020much, yadati2018link, yadati2019hypergcn}.
\vspace{-1.0\baselineskip}


\section{Conclusion}\label{conclusion}
In this paper, we study the problem of hypergraph representation learning, and propose an end-to-end hypergraph pre-training framework \petgnnh\ with GNNs. \petgnnh\ incorporates bi-level self-supervised pretext tasks (node-level and hyperedge-level respectively), and enables both transductive and inductive learning settings. The proposed pre-training strategy does not require extra domain-specific datasets, and is adaptation-aware, which helps the pre-trained model more adaptive to downstream tasks compared to traditional pre-training methods. Extensive experiments on public datasets as well as a real-world case study mainly demonstrate that: (1) \petgnnh\ shows significant improvements of downstream task performance and stability over baselines in most public datasets; (2) the proposed pre-training strategy is efficient compared with traditional pre-training method; (3) the proposed framework shows applicability in real-world applications of online stores.

\bibliographystyle{ACM-Reference-Format}
\bibliography{008reference}

\clearpage
\setcounter{secnumdepth}{0}
\section{Appendix}\label{sec:appendix}
We summarize the appendix as follows:
\begin{itemize}
    \item \textbf{Reproducibility:} The detailed experimental setups for training the proposed model, hyper-parameter setting, and dataset pre-processing for reproducibility.
    \item \textbf{Discussions of model variants:} The study of variants of the proposed \petgnnh.
    \item \textbf{Additional experimental results:} Additional results on model variants.
\end{itemize}

\subsection{A - Reproducibility}
\noindent \textbf{Model Training Details.} First, for \petgnnh\ with uniform negative sampling, we adopt a local sampling method for the negative hyperedges in order to sample negative nodes. Specifically, the uniform sampling is conducted inside each batch of hyperedges of the stochastic gradient descent algorithm (e.g. Adam optimizer). To this end, the node-level self-supervised pretext task captures the local node-context relationship within a batch at each parameter updating. Since the batch is uniformly sampled, this local sampling method eventually equals to the global uniform sampling on all hyperedges for pre-training. Compared with exponential sampling, which calculates $\mathbf{\tilde{A}}^k$ (Eq. \eqref{eq:prob}) for globally sampling negative nodes, this uniform negative sampling is more efficient (see Fig. \ref{fig:efficiency}).

Second, in order to obtain the hyperedge embeddings from the node embeddings as the outputs of GNN module, mean pooling is adpoted on public datasets and Set2set \cite{vinyals2015order} pooling with 1 recurrent layer is adopted on case study because of optimal practical performance. Other sophisticated pooling methods for hypergraphs could be one future direction. 

Third, all the adaptation functions which take the node/hyperedge embeddings as inputs for adapting with the self-supervised pretext/downstream tasks are set as MLPs with 0.5 dropout rate. All non-linear activation functions are set as ReLU function.

\noindent \textbf{Hyper-parameter Setting.}  For the model optimization, we adopt Adaptive Moment Estimation (Adam). The hyper-parameters are set such that the downstream task perform well on validation set. Specifically, for the optimizer, we select the batch size as 64, and learning rate as 0.001 with learning rate scheduler that reduces learning rate on plateau. The number of epochs for node-level pretext task and hyperedge-level pretext task (for traditional pre-training) are set to 50. For the effectiveness evaluations on public datasets, the number of clusters for hyperedge-level pretext task is set as $10, 10, 3, 3, 21, 20$ for \textit{Cora-noisy}, \textit{Cora-clean}, \textit{Pubmed-noisy}, \textit{Pubmed-clean}, \textit{Corum}, \textit{Disgenet} datasets, respectively. The number of GNN layers is set as 1. The number of adaptation steps are set to 5 for all datasets. The dimension of node/hyperedge embeddings are set as 64. 

For baseline methods, the hyper-parameters are empirically optimized based on the literature. For joint training, we set the weighting parameters $\alpha, \beta$ to be equal to 1 in Eq. \eqref{eq:joint}. For \textit{Hyper-SAGNN, DHNE, SAGE, and Joint Training} methods, we train them using Adam optimization method as well. The learning rate are set as 0.001 with learning rate scheduler. The dimension of node/hyperedge embeddings is set as 64. Other baseline hyper-parameters are set either based on the validation set or original literature guidance.

\noindent \textbf{Dataset Pre-processing.}\footnote{Note that if the hypergraph is constructed from regular graph, in which the \textit{pairwise} node links are known, the links among nodes inside and outside one hyperedge should neither be used for model input, nor be used for feature generation.} Since not all of the datasets are originally hypergraph datasets (i.e. \textit{Cora} and \textit{Pubmed}), we need to first process them for generating the hypergraphs. For \textit{Cora} and \textit{Pubmed}, we generate two versions of hypergraph datasets as follows. For the first version, we take the ego-network (subgraph of the center node with 1-hop neighbors) of each node as hyperedges, and assign the hyperedge label as the label of the majority in the ego-network. We name this version as the noisy version since the hyperedge might contain nodes with different categories. For the second version, we also take the ego-network of each node as hyperedges, but only keep those whose nodes share the same category. We name this version as the clean version. For \textit{Corum} and \textit{Disgenet}, they are originally hypergraph datasets, and do not need processing. 

As for the node features, for \textit{Cora} and \textit{Pubmed}, we use the text feature vectors as described in the dataset details. For \textit{Disgenet}, we use the numerical features of genes from the original data (e.g. DSI, DPI, etc.) For \textit{Corum}, we first construct a regular graph of nodes, in which each edge represents that the end nodes exist in the same hyperedge. Then we adopt the Subsample and Traverse (SaT) Walks \cite{payne2019deep} strategy for sampling a collection of random walks and use the embedding vectors from \textit{DeepWalk} \cite{perozzi2014deepwalk} method as node features\footnote{For all baselines (except for DeepWalk which does not utilize node features), we use the same node features as model inputs for fair comparison.}.
\hide{
\begin{figure}[b]
    \centering
    \includegraphics[width=0.45\textwidth, height=0.24\textwidth]{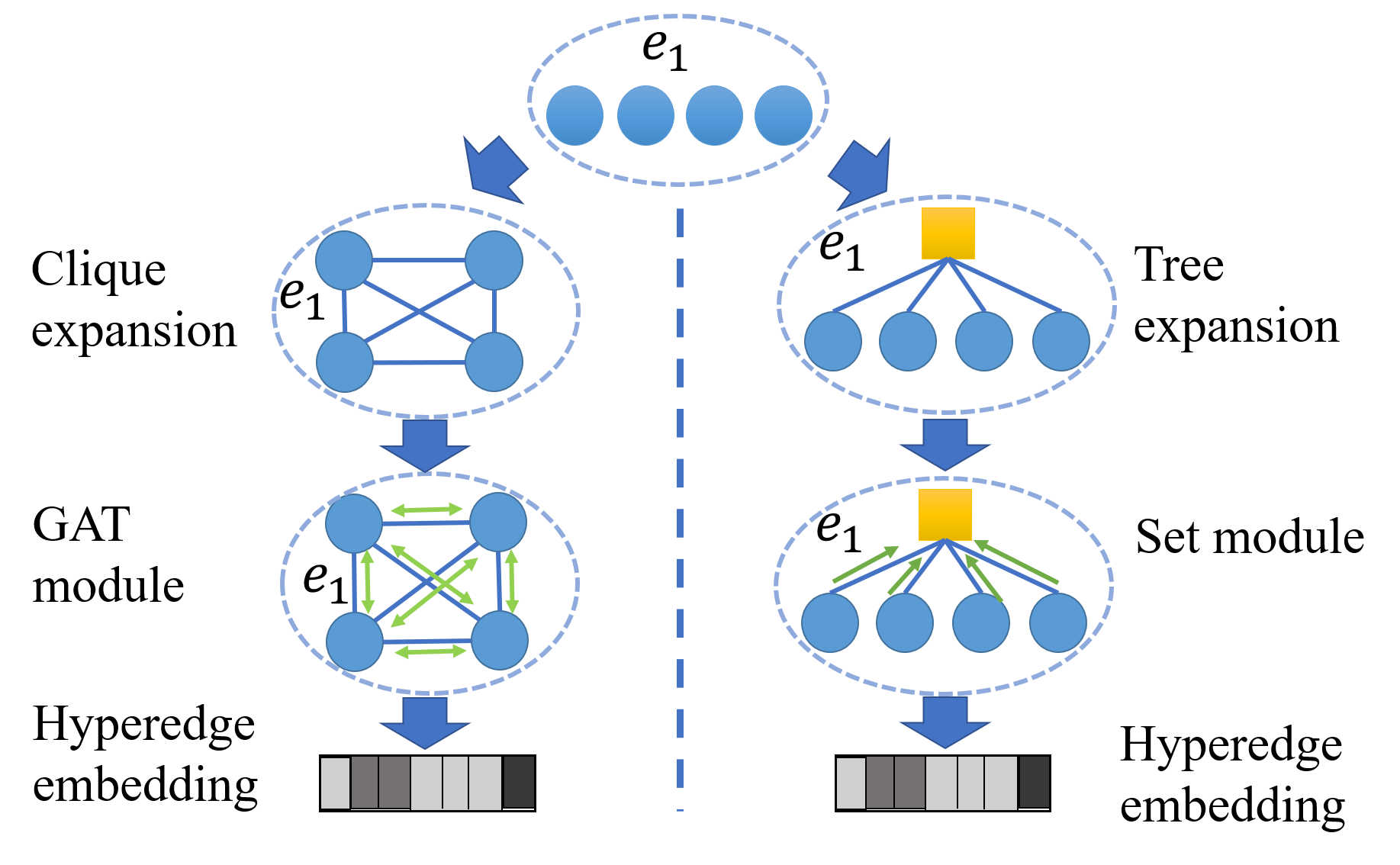}
    \vspace{-1.0\baselineskip}
    \caption{An illustrative example of the GNN module with clique expansion (left), and the tree expansion(right).}
    \label{fig:gnn}
\end{figure}
}

\begin{table*}
\caption{Comparison of transductive hyperedge classification performance for model variants}
\vspace{-1.0\baselineskip}
    \centering
    \begin{tabular}{l|p{20mm}|p{20mm}|p{20mm}|p{20mm}|p{20mm}|p{20mm}}
    \hline
         Models & Cora-noisy & Cora-clean & Pubmed-noisy & Pubmed-clean & Corum & Disgenet \\ \hline
         
         Variant-I & \textbf{78.97$\pm$3.69} & \textbf{84.13$\pm$2.65} & 84.72$\pm$1.22 & \underline{88.20$\pm$0.68} & \underline{84.65$\pm$0.05} & \underline{33.00$\pm$2.69}  \\
         
         Variant-II & 76.20$\pm$2.21 & 81.84$\pm$0.65 & 83.25$\pm$0.66 & 87.42$\pm$1.24 & 82.98$\pm$1.81 & 30.72$\pm$0.76  \\ 
      
         Variant-III &74.17$\pm$1.82 & 82.19$\pm$2,16 & 85.09$\pm$0.92 & 88.11$\pm$0.90 & 83.12$\pm$0.31 & 31.54$\pm$2.39 \\ 
         
         Variant-IV & 76.56$\pm$1.43 & 82.64$\pm$0.39 & \underline{85.56$\pm$0.39} & 87.32$\pm$0.97 & 68.24$\pm$1.41 & 32.15$\pm$2.10   \\
         
         \rowcolor{Gray}
         \petgnnh\ (Exp.) & \underline{78.78$\pm$1.28} & \underline{83.77$\pm$1.62} & \textbf{86.94$\pm$0.73} & \textbf{90.84$\pm$0.29} & \textbf{85.33$\pm$1.09} & \textbf{33.90$\pm$2.26}  \\ \hline
         
    \end{tabular}
    \label{tab:variant}
\end{table*}
\noindent \textbf{Baselines and Adjustments.} In total we adopt six baselines in our experiments. Five of them are from existing work (\textit{Deep Hyperedge} \cite{payne2019deep}, \textit{DHNE} \cite{tu2017structural}, \textit{Hyper-SAGNN} \cite{zhang2019hyper}, \textit{Graph-SAGE} \cite{hamilton2017inductive}, \textit{DeepWalk} \cite{perozzi2014deepwalk}), and one of them is the joint training strategy with the proposed self-supervised pretext tasks (Eq. \eqref{eq:joint}). 
Among the baselines from existing work, only \textit{Deep Hyperedge} is directly designed for hyperedge classification. \textit{DHNE} and \textit{Hyper-SAGNN} are originally designed for hyperlink prediction. We keep the major model architecture, and adapt these two methods as follows. 

For \textit{DHNE}, firstly the second-order component is designed for heterogeneous hyperedges. Since our datasets are not heterogeneous, we only need to use one auto-encoder in this component. Second, the supervised binary component for hyperlink prediction is modified to multi-class hyperedge classification. For \textit{Hyper-SAGNN}, we keep the idea of using both static and dynamic embeddings of nodes, and take the summation of the static and dynamic embeddings as the final embeddings of hyperedges for hyperedge classification. 

\textit{Graph-SAGE} and \textit{DeepWalk} are originally designed for regular graphs. We adapt these two models to make them work on the regular graph of hyperedges. In this regular graph, each vertex represents one hyperedge and there is an edge between two vertexes if two hyperedges share nodes. 

\noindent \textbf{Hardware and Software Details.} All experiments are performed on a machine with a Intel(R) Core(TM) i7-9800X CPU (64.0 GB RAM) and GeForce GTX 1080 GPU. The model is implemented with Python 3.6 and Pytorch 1.5.0 \cite{paszke2019pytorch}.

\subsection{B - Discussions of Model Variants}
\noindent \textbf{\petgnnh\ Algorithm.} We first present the proposed framework in Algorithm \ref{alg:framework}. 

\noindent \textbf{Joint Training.} Next we briefly describe a natural variant in which the pretext tasks are jointly trained with the downstream task. We also use this training method as a baseline in our experiments (see details in Section \ref{experiments}). The joint training loss can be written as:

\begin{equation}\label{eq:joint}
    \mathcal{L}^{(joint)}_{\Theta, \Omega_1, \Omega_2, \Omega_3} = \alpha\mathcal{L}^{(n)} + \beta\mathcal{L}^{(h)} + \mathcal{L}^{(t)}
\end{equation}
where $\mathcal{L}^{(t)}$ represents the loss function for downstream task, hyperedge classification. $\alpha, \beta$ are used for re-weighting the bi-level pretext tasks. 

\begin{algorithm}
    \SetKwFunction{isOddNumber}{isOddNumber}
    \SetKwInOut{KwIn}{Input}
    \SetKwInOut{KwOut}{Output}
    
    \KwIn{$\mathcal{E}_{train} = \{e_1, ..., e_m\}$, $\mathcal{E}_{test} = \{e_1, ..., e_q\}$: training and testing set for hyperedge classification; pretext tasks' hyper-parameters; }
    \KwOut{The pre-trained GNN module $f_\Theta(\cdot)$.}

    \For{$e \in \mathcal{E}_{train}\cup\mathcal{E}_{test}$}{
        Perform clique expansion for $e$ to obtain new hyperedge $e' \in \mathcal{E'}_{train}\cup\mathcal{E'}_{test}$.

    }
    \For{each epoch}{
        Sample batches $\mathcal{B}^{(m)} = \{\mathcal{B}_1, \mathcal{B}_2, ..., \mathcal{B}_s\}$ from $\mathcal{E'}_{train}$. 
        
        \For{each batch $\mathcal{B}_i \in \mathcal{B}^{(m)}$}{
            Update GNN module parameters $\Theta$ by $\mathcal{L}^{(n)}$ (Eq. \eqref{eq:loss_node}).
        }
    }
    \tcc{Line 11 - 16: for inductive setting only}
    \For{each epoch}{
        Sample batch set $\mathcal{B}^{(m)} = \{\mathcal{B}_1, \mathcal{B}_2, ..., \mathcal{B}_s\}$ from $\mathcal{E'}_{train}$. 
        
        \For{each batch $\mathcal{B}_i \in \mathcal{B}^{(m)}$}{
            Update GNN module parameters $\Theta$ by $\mathcal{L}^{(h)}$ (Eq. \eqref{eq:loss_he}).
        }
    }
    
    \tcc{Line 18 - 20: for transductive setting only}
    \For{each adaptation step}{
         Update GNN module parameters $\Theta$ by Eq. \eqref{eq:ada_step} on $\mathcal{E'}_{test}$.
    }
   
    Return Pre-trained GNN module parameter $\Theta$.
    \caption{The \petgnnh\ framework}
    \label{alg:framework}
\end{algorithm}

\noindent \textbf{Variant I.} As discussed in Section \ref{sec:architecture}, instead of using one GNN module for both positive and negative nodes in the node-level pretext task, one natural variant is to utilize two GNN modules, which do not share parameters, for positive and negative nodes respectively. 

\noindent \textbf{Variant II.} For the node-level self-pretext task, we could also learn the hyperedge representation by a ranking-based objective function, which aims at forcing the similar nodes in the feature space to be also close in the representation space, and the dissimilar nodes to having a margin in the representation space. Here we adopt the cosine embedding loss function to replace Eq. \eqref{eq:loss_node}. For node $i$ and context $j$: 

\begin{equation}
    \mathcal{L}^{(n)} = \sum_{i,j} (y(1 - cos(\mathbf{\hat{h}}_i, \mathbf{\hat{h}}^{(C)}_j)) + (-y)(max(0, cos(\mathbf{\hat{h}}_i, \mathbf{\hat{h}}^{(C)}_j) - \epsilon))
\end{equation}
where $\epsilon>0$ is a margin, $y\in\{-1,1\}$ is the label, and $cos(\cdot)$ is the cosine similarity function.

\noindent \textbf{Variant III.} For the hyperedge-level self-pretext task, in order to approximate the task of preserving the exact pair-wise hyperedge similarities, we can also try to preserve the approximated pair-wise hyperedge distances in a regression objective as follows to replace Eq. \eqref{eq:loss_he}.
\begin{equation}
    \mathcal{L}^{(h)} = ||g^{(h)}_{\Omega_2}(\mathbf{H})^{\T}g^{(h)}_{\Omega_2}(\mathbf{H}) - \mathbf{A}^{k'}||_F^2
\end{equation}
where $g^{(h)}_{\Omega_2}(\cdot)$ is a neural adjustment function for the embeddings of hyperedges, $\mathbf{H}$ is the embedding matrix of all hyperedges (with row-wise hyperedge embeddings), $k'>0$ is a small scalar and $\mathbf{A}^{k'}$ is an approximation for pair-wise hyperedge similarities when the incidence matrix can be used for calculating the adjacency matrix $\mathbf{A}$. In this Variant, we use the above pretext task as adaptation steps.

\noindent \textbf{Variant IV.}
In order to adopt GNN technique and learn the representations for hyperedges. Besides expanding each hyperedge as a fully-connected graph ({\em clique expansion}), we can also expand a hyperedge as a tree ({\em tree expansion}). Intuitively, each hyperedge can be seen as a new root node connecting to all the nodes inside the hyperedge. In the message passing of GNN module, the difference between clique expansion and tree expansion is that the nodes in clique expansion first pass their features to the neighbors in the same hyperedge, but the nodes in tree expansion only pass their features to the root nodes. Clique expansion and tree expansion are two typical methods of adding hypothetical connections to the nodes inside hyperedges for expansion.

\subsection{C - Additional Experimental Results}
We present the experimental results on the four variants which we discuss in the Appendix-B in Table. \ref{tab:variant}. Compared with \petgnnh\ with exponential negative sampling, the Variant-I with two GNN modules in transductive setting achieves slightly better performance in \textit{Cora} dataset. Variant-II and Variant-III with different node-level/hyperedge-level pretext task objectives could not improve the performance of \petgnnh. Variant-IV with tree expansion could also not achieve improvements over original methods. Further sophisticated model architectures and variants are left for future work.

\end{document}